\documentclass{article}

\usepackage[final]{neurips_2025}

\usepackage[utf8]{inputenc} % allow utf-8 input
\usepackage[T1]{fontenc}    % use 8-bit T1 fonts
\usepackage{hyperref}       % hyperlinks
\usepackage{url}            % simple URL typesetting
\usepackage{booktabs}       % professional-quality tables
\usepackage{amsfonts}       % blackboard math symbols
\usepackage{nicefrac}       % compact symbols for 1/2, etc.
\usepackage{microtype}      % microtypography
\usepackage{xcolor}         % colors

\usepackage{graphics}
\usepackage{float}
\usepackage{epsfig} 
\usepackage{mathptmx}
\usepackage{times} 
\usepackage{amsmath}
\usepackage{amssymb} 
\usepackage{pifont}

\usepackage{multirow}
\usepackage{colortbl}  %彩色表格需要加载的宏包
\usepackage{xcolor}
\usepackage{array}   %对表列和表格线的设置需要用到array宏包
\usepackage{makecell}       % for \makecell in tables
\usepackage{bm}
\usepackage{enumitem}
\usepackage{caption}
\usepackage{graphicx}
\usepackage{ulem}

\begin{document}

\author{
  Weijie Zhou$^{1,2,3,\ast}$ \hspace{0.2cm} 
  Xuantang Xiong$^{2,\ast}$ \hspace{0.2cm}
  Yi Peng$^{3}$ \hspace{0.2cm}
  Manli Tao$^{3}$ \\
  \textbf{Chaoyang Zhao}$^{3,4,\dagger}$ \hspace{0.2cm}
  \textbf{Honghui Dong}$^{1}$ \hspace{0.2cm}
  \textbf{Ming Tang}$^{3}$ \hspace{0.2cm}
  \textbf{Jinqiao Wang}$^{3,4,\dagger}$ \\
  $^1$Beijing Jiaotong University \\ 
  $^2$Tencent Robotics X \& Futian Laboratory, Shenzhen \\ 
  $^3$Foundation Model Research Center, Institute of Automation, Chinese Academy of Sciences \\
  $^4$ObjectEye Inc.
}

\footnotetext[1]{Equal contribution}
\footnotetext[2]{Corresponding authors}

\title{PhysVLM-AVR: Active Visual Reasoning for Multimodal Large Language Models in Physical Environments}
\maketitle

\begin{abstract}

Visual reasoning in multimodal large language models (MLLMs) has primarily been studied in static, fully observable settings, limiting their effectiveness in real-world environments where information is often incomplete due to occlusion or limited field of view. Humans, in contrast, actively explore and interact with their environment—moving, examining, and manipulating objects—to gather information through a closed-loop process integrating perception, reasoning, and action. Inspired by this human capability, we introduce the \textbf{Active Visual Reasoning (AVR)} task, extending visual reasoning to partially observable, interactive environments. AVR necessitates agents to: (1) actively acquire information via sequential physical actions, (2) integrate observations across multiple steps for coherent reasoning, and (3) dynamically adjust decisions based on evolving visual feedback. To rigorously evaluate AVR, we introduce \textbf{CLEVR-AVR}, a simulation benchmark featuring multi-round interactive environments designed to assess both reasoning correctness and information-gathering efficiency. We present \textbf{AVR-152k}, a large-scale dataset offers rich Chain-of-Thought (CoT) annotations detailing iterative reasoning for uncertainty identification, action-conditioned information gain prediction, and information-maximizing action selection, crucial for training agents in a higher-order Markov Decision Process. Building on this, we develop \textbf{PhysVLM-AVR}, an MLLM achieving state-of-the-art performance on CLEVR-AVR, embodied reasoning (OpenEQA, RoboVQA), and passive visual reasoning (GeoMath, Geometry30K). Our analysis also reveals that current embodied MLLMs, despite detecting information incompleteness, struggle to actively acquire and integrate new information through interaction, highlighting a fundamental gap in active reasoning capabilities.
\end{abstract}

\section{Introduction}

\begin{figure*}[ht] % [H]
    \begin{center}
    \centerline{\includegraphics[width=1\columnwidth]{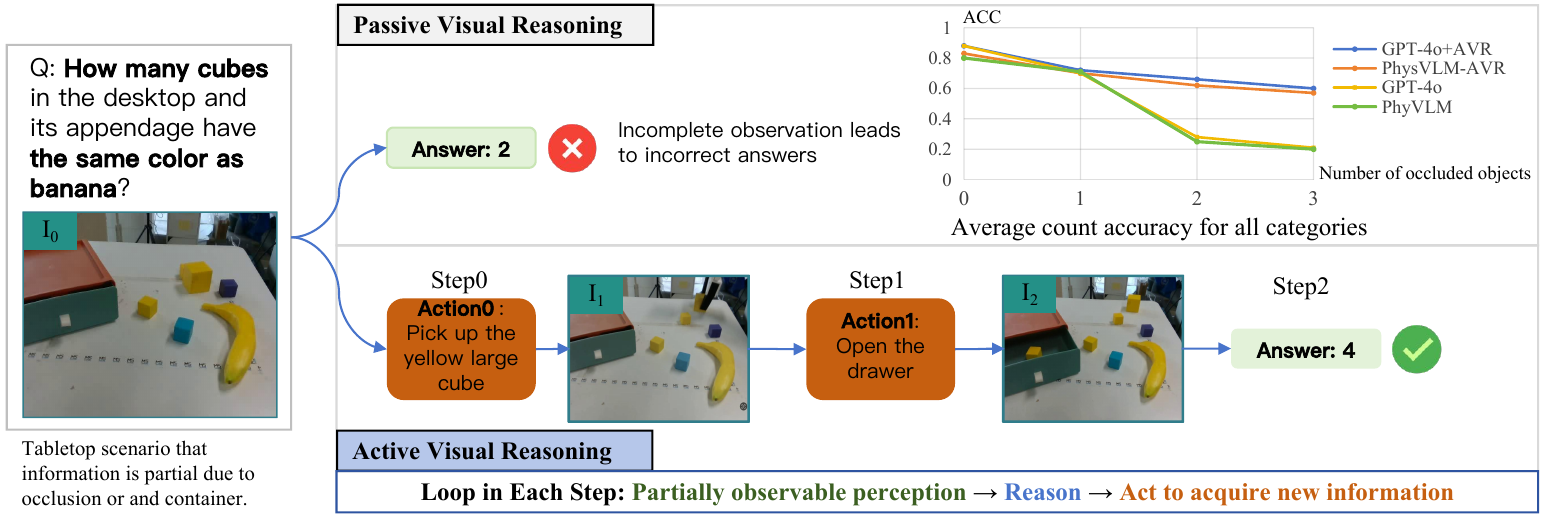}}
    \caption{Passive Visual Reasoning (top) fails with partial views. Active Visual Reasoning (AVR, bottom) actively interacts to gather information for a correct answer.}
    \label{image1}
    \vspace{-0.5cm}
    \end{center}
\end{figure*}

Visual reasoning, a fundamental capability in artificial intelligence, has enabled multimodal large language models (MLLMs) \cite{qwen2vl, llava, llavanext, internvl, claude3} to excel at tasks such as object counting \cite{CLEVR, cleve-math, SuperCLEVR} and visual question answering (VQA) \cite{Visionr1, VisionR1zhan, Visualprogramming, vqav2}. These abilities are crucial for developing intelligent systems that can comprehend and interact with the visual world. As MLLMs continue to evolve, their potential to understand and reason about dynamic, real-world environments is increasingly recognized, paving the way for more sophisticated and autonomous agents \cite{robobrain, roboos, physvlm, physical1}.

Despite these advancements, current MLLM approaches to visual reasoning—including prominent tasks like VQA \cite{okvqa, vqav2, eqa1, eqa2} and spatial reasoning \cite{spatialvlm, spatialbot, geminirobotics}—predominantly rely on static, passive visual inputs. This inherently limits their applicability in real-world physical environments where information is often partially observable due to occlusions, containment, or limited field of view \cite{physical2}. As illustrated in Figure~\ref{image1}, a passive model observing only the initial image $I_0$ answers incorrectly, whereas an active agent interacts to uncover occluded objects across $I_1$ and $I_2$, arriving at the correct answer. And in the task of counting all object categories in the scene, the more occluded objects there are, the worse the model performs in passive visual reasoning mode. This highlights that effective reasoning in partially observable settings necessitates interaction to actively acquire missing information.

Humans, in stark contrast, naturally overcome such limitations through active perception. We instinctively move, change viewpoints, and manipulate objects to gather information needed for a specific goal. This active engagement exemplifies a core aspect of human cognition: \textit{closing the perception-reasoning-action loop to incrementally resolve uncertainties and establish the causal links essential for complex reasoning.}

However, existing computational frameworks often suffer from paradigm fragmentation and thus fail to holistically model the complete perception-reasoning-action loop. Conventional visual reasoning tasks \cite{Reasonrft, Visionr1, VisionR1zhan} (e.g., CLEVR \cite{CLEVR}) assume complete visual inputs, requiring only a single "perception → reasoning" step without active information gathering \cite{mmbench, mmvet, mmebenchmark}. While embodied question answering (EQA) tasks (e.g., OpenEQA \cite{openeqa}, RoboVQA \cite{robovqa}) involve interaction, they often focus on reasoning from single, albeit potentially long, video sequences. Similarly, embodied exploration methods \cite{askforhelp, embodiedReasoner, Knowledge-based-eqa}, also interactive, tend to prioritize task completion metrics (e.g., navigation success) over the explicit information gain needed for nuanced reasoning. The fundamental limitation across these approaches is their failure to effectively link reasoning with the strategic, sequential actions essential for information gathering. Specifically, they often overlook how an agent's actions dynamically alter available information and, crucially, how these changes should guide and refine the ongoing reasoning process, especially when multiple evidence-gathering steps are required.

Inspired by human active perception and to address these limitations, we introduce the \textbf{Active Visual Reasoning (AVR)} task. AVR extends visual reasoning to partially observable, interactive environments, bridging the gap between passive observation and active, embodied understanding. Specifically, AVR combines embodied interaction with temporal visual reasoning, requiring agents to: (1) actively gather information through sequential physical actions; (2) integrate multi-step observations for coherent reasoning; and (3) dynamically adjust decisions based on incremental visual feedback. This paradigm requires models to interpret visual data and strategically decide how to interact with the environment, thereby resolving uncertainties crucial for complex reasoning.

To establish a rigorous foundation for the AVR task and facilitate research in this domain, this paper makes the following primary contributions:

\begin{itemize}[leftmargin=2em]
\item We formally introduce and define the \textbf{Active Visual Reasoning (AVR)} task, mandating active information gathering, multi-step integration, and dynamic decision-making in partially observable settings.
\item We develop \textbf{CLEVR-AVR}, a simulation benchmark featuring multi-round interactive environments. This benchmark is specifically designed to assess both the reasoning correctness and the information-gathering efficiency of agents performing AVR tasks.
\item We present \textbf{AVR-152k}, a large-scale dataset with multi-level annotations designed to train agents for AVR. Its core component, AVR-Core, models the task as a higher-order Markov Decision Process and features rich Chain-of-Thought (CoT) annotations. These CoTs articulate the structured, iterative reasoning humans employ for active information seeking, including critical steps such as: (i) identifying information uncertainty, (ii) predicting action-conditioned information gain, and (iii) selecting information-maximizing actions, thereby providing explicit supervision for these decision-making processes.
\item We develop and evaluate \textbf{PhysVLM-AVR}, an MLLM that achieves state-of-the-art performance on CLEVR-AVR while maintaining strong results on standard embodied visual reasoning tasks, and reveals insights into current MLLM limitations regarding active interaction.
\end{itemize}

Collectively, our work provides a foundation for developing MLLMs capable of actively reasoning about and intelligently interacting with their physical environments, thereby narrowing the gap between static visual reasoning and embodied intelligence.

\section{Related Work}

\textbf{Visual Reasoning.} Traditional visual reasoning tasks, such as Insight-V \cite{insightv}, CLEVR \cite{CLEVR} and its variants (e.g., CLEVR-Math \cite{cleve-math}, MathVision \cite{awais2024mathvision}), have significantly advanced models' abilities in attribute recognition, counting, and spatial reasoning \cite{robospatial}. However, these tasks operate on static, fully observable images, lacking mechanisms for active information seeking in partially observable scenarios. Consequently, they do not model the sequential, interactive process essential for real-world understanding. AVR addresses this by requiring agents to perform sequential actions to actively acquire information necessary for reasoning.

\textbf{Embodied Question Answering.} Embodied Question Answering (EQA) tasks (e.g., OpenEQA \cite{openeqa}, RoboVQA \cite{robovqa}, CityEQA \cite{cityeqa}) extend question answering to simulated embodied environments. However, by often relying on pre-captured data, these approaches limit agents to passive observation rather than active, dynamic exploration to resolve current uncertainties. As a result, the critical closed-loop interaction – where reasoning about the question drives information-seeking actions, and new information from these actions refines understanding – is frequently underemphasized. AVR, in contrast, centers on this loop, compelling agents to integrate multi-step observations acquired through their own goal-directed actions.

\textbf{Embodied Exploration.} Embodied exploration and task planning methods (e.g., Knowledge-based EQA \cite{Knowledge-based-eqa}, Embodied Reasoner \cite{embodiedReasoner}, ET-Plan-Bench \cite{etPlanBench}, EXPRESS-Bench \cite{expressBench}) enable agents to interact with their environments for broader goals like navigation or multi-step task completion. However, these approaches typically optimize for overall task success (e.g., navigation efficiency) rather than the targeted information acquisition crucial for active reasoning. Their evaluation often prioritizes general task metrics over the specific information gain pertinent to a reasoning objective. AVR, conversely, prioritizes dynamic, uncertainty-driven action selection to maximize information gain directly relevant to the goal, highlighting the link between information-seeking actions and reasoning refinement.

In summary, while existing research touches upon interaction and reasoning, AVR uniquely integrates these by emphasizing active perception driven by reasoning needs. This directly addresses robust understanding in partially observable environments, mirroring a key aspect of human cognition.

\section{Active Visual Reasoning}

\subsection{Problem Formulation}

We define \textbf{Active Visual Reasoning (AVR)} as a closed-loop paradigm where an agent must actively interact with a partially observable environment $E$ using a finite set of atomic actions ${A}$ to answer a question $Q$. Unlike conventional visual reasoning, AVR models reasoning as an iterative perception-reasoning-action cycle.
At each timestep $t\in \{0,1,...,T_{\text{max}}\}$, the agent receives a partial observation $o_t$ and maintains an observation history $h_t = \{o_0,...,o_t\}$. Based on $Q$ and $h_t$, the reasoning module $f_{\text{reason}}$ generates an intermediate reasoning trace:
\begin{equation}
    \text{Think}_t = f_{\text{reason}}(Q, h_t, {A}).
\end{equation}

The agent then assesses if $h_t$ is sufficient to answer $Q$. If so, it generates an answer $y_t = f_{\text{answer}} (Q, h_t)$. Otherwise, it selects an optimal information-gathering action $a_t$ to maximize expected information gain about the true answer $Y$:
\begin{equation}
    a_t = \text{argmax}_{a_t\in A} \mathbb{E}_{o_{t+1} \sim E(h_t, a_t)} \left[ I(Y;y_{t+1}|h_{t+1},Q) \right], \label{eq:action}
\end{equation}

where $h_{t+1} = (h_t, o_{t+1})$. Executing $a_t$ yields $o_{t+1}$, and $h_{t+1}$ is updated. This cycle iterates until an answer is generated or $T_{\text{max}}$ is reached.

AVR thus integrates three key components: (1) \textbf{Active Information Acquisition}: Strategically interacting to gain relevant information. (2) \textbf{Temporal Visual Integration}: Reasoning over accumulated sequential observations. (3) \textbf{Dynamic Decision-Making}: Continuously adapting beliefs and actions based on new evidence.

\subsection{Key Challenges}

AVR presents distinct challenges beyond static visual reasoning: (1) \textbf{Efficient Exploration}: Strategically exploring partially observable environments to acquire task-relevant information with minimal actions. (2) \textbf{Temporal Visual Reasoning}: Integrating and reasoning over extended observation sequences from multi-step interactions. (3) \textbf{Reasoning-Driven Actions}: Ensuring actions are guided by current understanding to purposefully reduce uncertainty, avoiding random exploration.

Addressing these multifaceted challenges necessitates dedicated benchmarks and datasets. Accordingly, the CLEVR-AVR benchmark and the AVR-152k dataset, detailed in the next section, are designed to encapsulate these difficulties. Our dataset construction, in particular, captures the sequential decision-making processes inherent in AVR, providing a structured foundation for developing and evaluating agents for these complex tasks.

\section{CLEVR-AVR Benchmark and AVR-152k Dataset}
\label{sec:benchmark_dataset}

\subsection{CLEVR-AVR: A Benchmark for Evaluating AVR Capabilities}
\label{subsec:clevr_avr}

The \textbf{CLEVR-AVR} benchmark is a simulation-based evaluation suite designed to rigorously assess an agent's proficiency in the AVR paradigm. Leveraging the Genesis \cite{Genesis} physical simulation platform, it extends the classic CLEVR \cite{CLEVR} setup into an interactive, embodied domain. This provides a controlled yet rich environment for studying the closed loop of perception, reasoning, and action fundamental to AVR. Further details on the scene types, question template structures, and the distribution of occlusion and stacking challenges within CLEVR-AVR are provided in Appendix Figures~\ref{fig:scene_examples}, \ref{fig:question_templates}, and \ref{fig:occlusion_distribution}, respectively.

\begin{figure*}[ht] % [H]
    \begin{center}
    \centerline{\includegraphics[width=1\columnwidth]{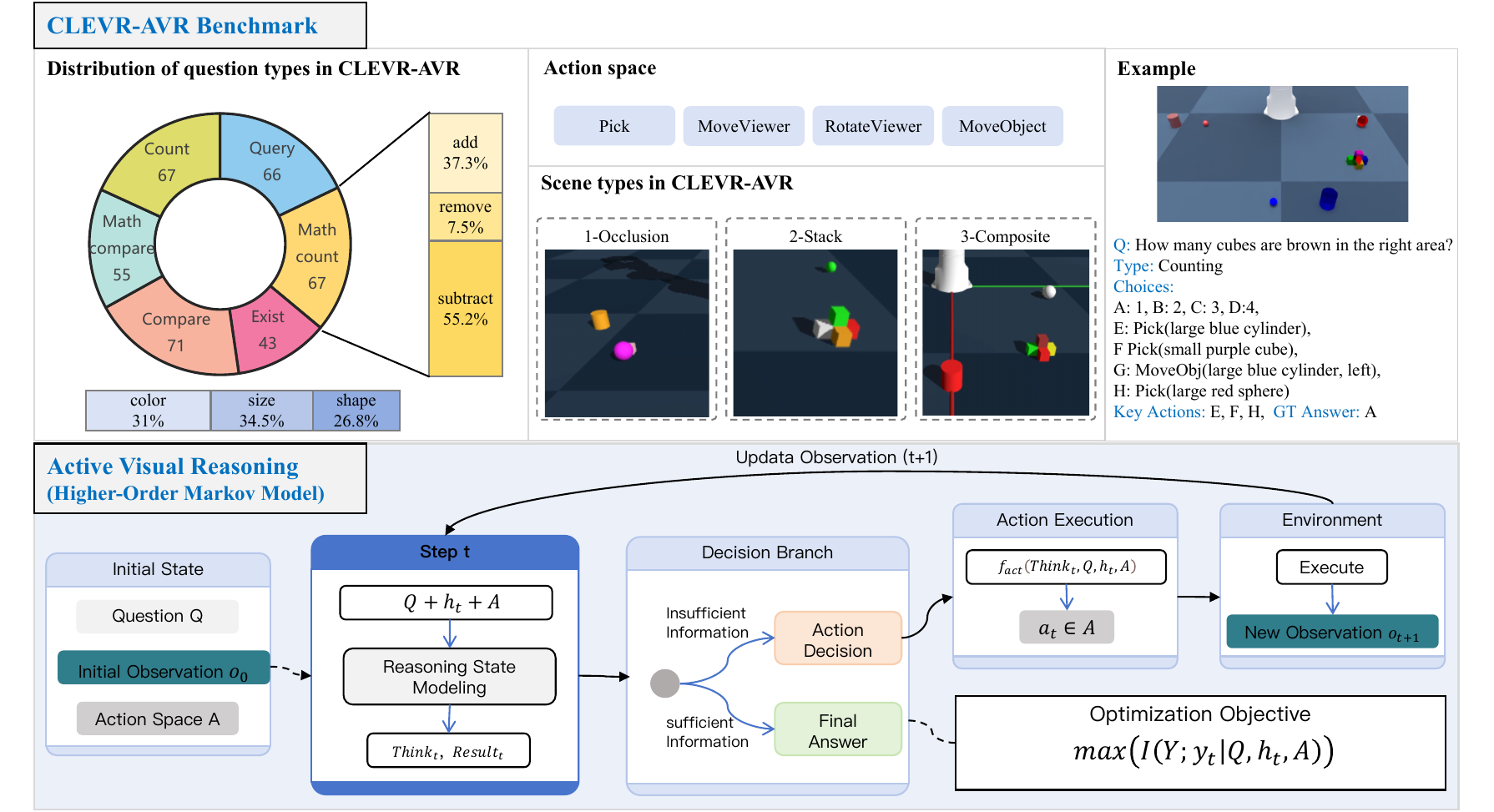}}
    \caption{Top: CLEVR-AVR Simulator Benchmark, showing distributions of question types, action space, scenes, and examples. Bottom: Higher-order Markov Decision Process (MDP) paradigm for Active Visual Reasoning (AVR).}
    \label{image2}
    \vspace{-0.5cm}
    \end{center}
\end{figure*}

\textbf{Challenge of Efficient Exploration in Partial Observability.}
A core tenet of AVR is overcoming incomplete information. As shown in Figure \ref{image2}, CLEVR-AVR confronts agents with diverse scenarios featuring 10 occlusion types, 10 stacking types, and 10 composite scenarios combining these elements. The action space includes object manipulation and viewpoint changes, with agents required to interactively gather information to answer diverse question types. For detailed action formats and candidate generation, see Appendix Sec.~\ref{appendix:action_generation_details}.

\textbf{Challenge of Temporal Visual Reasoning.}
CLEVR-AVR demands multi-step integration of observations for coherent reasoning. On average, each question requires at least two reasoning steps to resolve, with some scenarios designed to necessitate 4-6 interaction steps.

\textbf{Challenge of Reasoning-Driven Actions.} 
Each question in CLEVR-AVR is paired with candidate responses that include both final answer options and intermediate \texttt{[Action]} options, as exemplified in Figure \ref{image2}. This structure allows agents to either provide a conclusive answer or explicitly choose to interact further with the environment. Agents must assess whether current information is sufficient or if an action is necessary to reduce uncertainty, promoting purposeful, uncertainty-reducing interactions over random exploration. (Further details on action candidate generation are in Appendix Sec.~\ref{appendix:action_generation_details}).

\textbf{Evaluation Metrics.}
\label{Metrics}

CLEVR-AVR primarily evaluates: (1) \textbf{Information Sufficiency Judgment Accuracy ($ACC_{ISJ}$):} Accuracy in judging whether the initial observation provides sufficient information to answer the question. For example, if crucial objects for a counting task are occluded in the initial view, a correct judgment would be to acknowledge insufficiency and select an action option rather than attempting a premature final answer. (2) \textbf{Information Gain Rate ($IGR$):} This metric is derived by dividing the count of information-gaining decisions by the total number of decision steps. 
(3) \textbf{Final Answer Accuracy ($ACC_{FA}$):} Correctness of the final answer, with ground truth from the simulator.

An action achieves \textit{Information Gain} if it reveals new, question-relevant visual information (e.g., uncovering hidden objects, changing viewpoints for obscured areas). This is determined by simulator ground truth: an action gains information if it unhides relevant objects/surfaces previously occluded or stacked upon. This objectively measures if the action yielded a more complete observation, crucial for tasks like counting or attribute querying needing comprehensive exploration. These metrics together comprehensively assess an agent's efficient and accurate Active Visual Reasoning (AVR).

\subsection{AVR-152k Dataset: Modeling Active Visual Reasoning as a Higher-Order MDP}
\label{subsec:avr_mdp_data}

To facilitate the development of agents capable of Active Visual Reasoning (AVR)—particularly in addressing the challenges of \textbf{Efficient Exploration, Temporal Visual Reasoning, and Reasoning-Driven Actions}—we introduce the \textbf{AVR-152k dataset}. A key component of this dataset, \textbf{AVR-Core}, is specifically designed to model AVR tasks as a \textbf{higher-order Markov Decision Process (MDP)}, as illustrated in Figure~\ref{image2}. This MDP formulation provides a principled framework for agents to learn how to sequentially gather information and reason in interactive environments.

\begin{figure*}[ht] % [H]
    \begin{center}
    \centerline{\includegraphics[width=0.97\columnwidth]{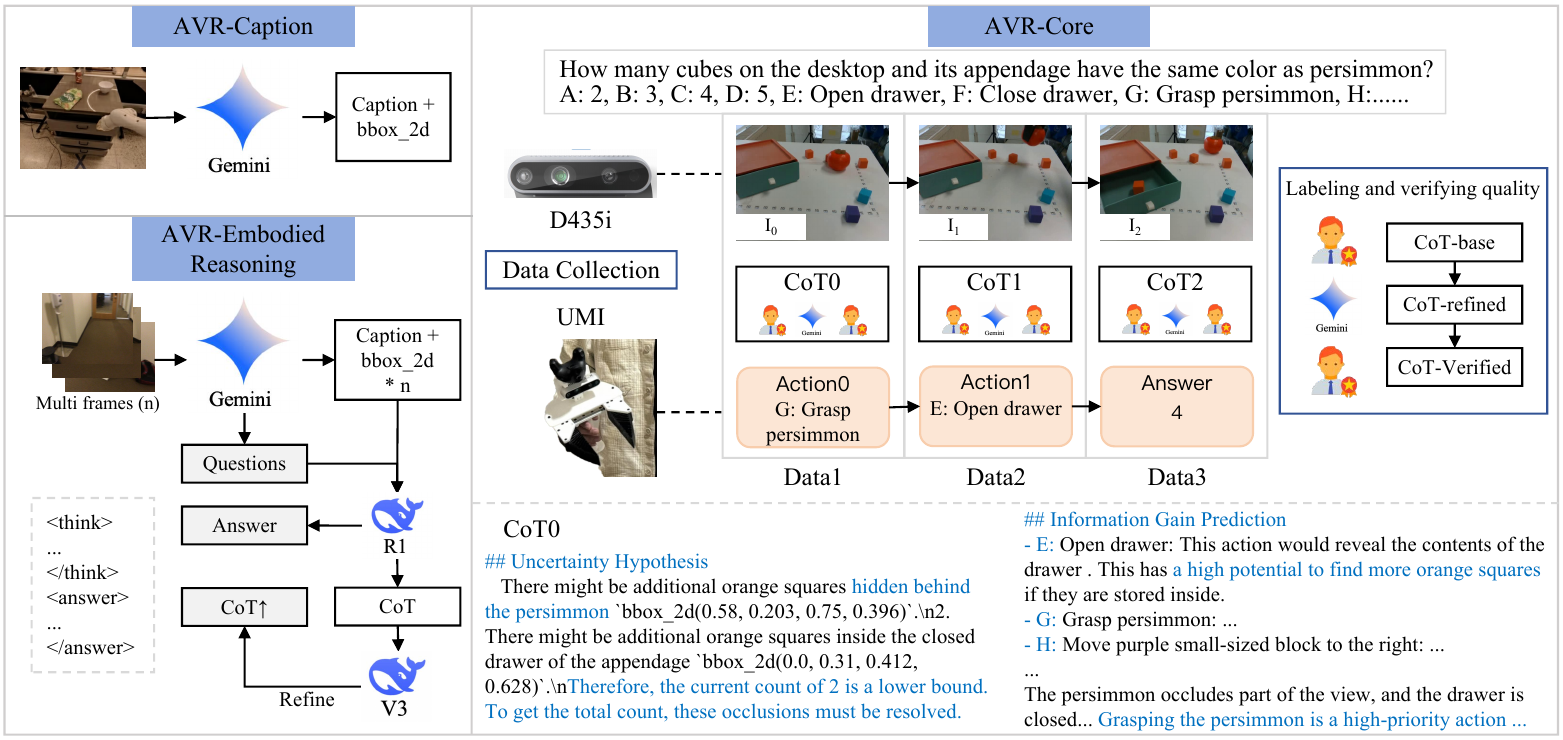}}
    \caption{AVR-152k Dataset Construction. Left: Workflows for AVR-Caption and AVR-Embodied Reasoning (perception, temporal reasoning). Right: AVR-Core details including its sequential annotation, CoT supervision, and quality verification.}
    \label{image3}
    \vspace{-0.6cm}
    \end{center}
\end{figure*}

Within this MDP, As shown in Figure~\ref{image3}, AVR-Core provides rich Chain-of-Thought (CoT) annotations for the reasoning state ($\text{Think}_t$). These CoTs are meticulously structured to reflect a natural, human-like information-seeking process: (1) \textit{Assessing Current Understanding}: evaluating the information available from observation history ($h_t$) in relation to the question ($Q$) and identifying key uncertainties or missing details. (2) \textit{Evaluating Potential Actions}: considering available actions ($A$) and forecasting their potential to resolve identified uncertainties and yield valuable new information. (3) \textit{Strategic Decision-Making}: based on the assessment and evaluation, deciding whether to take an information-gathering action ($a_t$) or, if uncertainty is sufficiently resolved, provide a final answer.
Training with these explicit CoT annotations enables models to internalize this multi-step reasoning. 

The full AVR-152k dataset, including its foundational subsets, supports training models to master this complex, interactive reasoning loop. As shown in Figure \ref{image3}, AVR-152k comprises three subsets with progressively increasing complexity:

\textit{AVR-Caption (100k samples):} Focuses on foundational visual perception and spatial understanding, providing dense captions with bounding boxes for indoor scenes from diverse embodied datasets (e.g., ScanNet \cite{scannet}, RT1 \cite{rt1}). Captions were generated by Gemini-2.0-flash \cite{gemini} using the prompt detailed in Appendix Figure~\ref{fig:caption_prompt}, and an example data instance is shown in Appendix Figure~\ref{fig:data-caption}.

\textit{AVR-Embodied Reasoning (50k samples):} Advances to \textbf{temporal visual reasoning}. It consists of multi-image sequences paired with questions requiring spatiotemporal understanding. Dense captions were generated by Gemini-2.0-flash, while DeepSeek-R1-671B \cite{deepseekr1} produced reasoning chains (see Appendix Figure~\ref{fig:embodied_reasoning_prompt} for the prompt), which were subsequently optimized by DeepSeek-V3 \cite{deepseekv3} (see Appendix Figure~\ref{fig:cot_refine_prompt} for the refinement prompt). An illustrative example from this subset is provided in Appendix Figure~\ref{fig:data-embodied-reasoning}.

\textit{AVR-Core (2k samples):} Directly instantiates the higher-order MDP for AVR. Data was collected using UMI \cite{umi} devices interacting with 640 real-world tabletop settings. As shown in Figure~\ref{image3}, each sample in AVR-Core contains expert-structured CoT annotations for the $\text{Think}_t$ state, initially authored by human experts through a multi-step process (details in Appendix Sec.~\ref{appendix:cot_generation_detail}), and subsequently refined by Gemini (see Appendix Figure~\ref{fig:gemini_refine_prompt} for the refinement prompt). These annotations explicitly demonstrate the human-like reasoning process. They detail how to assess uncertainty, predict information gain from potential actions, and articulate the rationale for selecting an action or providing a final answer. This iterative process is exemplified in the multi-step interactive reasoning sequences shown in Appendix Figures~\ref{fig:data-core-step0}, \ref{fig:data-core-step1}, and \ref{fig:data-core-step2}. Questions typically involve several reasoning-and-action steps (avg. 3.2), requiring models to repeatedly apply this iterative decision-making process. AVR-Core underwent rigorous validation: expert annotation, logical consistency verification by Gemini \cite{gemini}, and human expert review of perception-reasoning-action validity.

The higher-order MDP embodied by AVR-Core (illustrated in Figure~\ref{fig:data-core-step0}, \ref{fig:data-core-step1}, and \ref{fig:data-core-step2}) formalizes this iterative, closed-loop process. Each sample in AVR-Core captures a single step of this interaction. It provides the current context, including the question $Q$, visual observation history $h_t$, past action history $a_{0:t-1}$, and the set of available actions $A$. Crucially, it contains the expert-annotated reasoning trace $\text{Think}_t$, which details the assessment of current understanding, evaluation of potential actions, and strategic decision-making. As an approximation of the goal in Equation~\ref{eq:action}, This reasoning culminates in a recorded outcome: either a chosen information-gathering action $a_t \in A$ or a final answer $y_t$. If an action $a_t$ is selected, the resulting new visual observation $o_{t+1}$ from the environment $E$ is also included, forming the basis for the next step. This structured data allows models to learn the mapping from the current state and available actions to the generation of a reasoning trace $\text{Think}_t$ and the subsequent decision.

AVR-152k, with its structured CoT annotations within an MDP framework, offers a principled approach for developing models that actively seek information for reasoning in interactive environments.

\section{PhysVLM-AVR Model}
\label{sec:physvlm_avr}

To equip a MLLM with active visual reasoning capabilities, we develop \textbf{PhysVLM-AVR}. The architecture of PhysVLM-AVR is similar to LLaVA \cite{llava}, employing Qwen2.5-3B \cite{qwen2.5} as the LLM decoder and SigLIP-400M \cite{siglip} as the visual encoder. A key modification for efficient multi-image reasoning is the introduction of a max pooling layer after the visual encoder's output, reducing the number of visual tokens by a factor of three. This allows for more efficient processing of multiple visual inputs. More details of the model architecture see Appendix~\ref{Appendix-model-architecture}.

We employ a multi-stage, mixed-data training strategy to progressively build the generalizable active visual reasoning capabilities of PhysVLM-AVR:

\begin{itemize}[leftmargin=2em]
    \item \textbf{Stage 1: Alignment.} In this initial stage, we focus on aligning visual features with the language model. We train only the connector layers (2xMLP) using the LLaVA-Pretrain \cite{llava} dataset.
    \item \textbf{Stage 2.1: Single-Image Understanding.} To develop foundational image comprehension, all  parameters are fine-tuned using the LLaVA-OneVision-data \cite{llavanext}.
    \item \textbf{Stage 2.2: Comprehensive Visual Understanding.} We then enhance the model's broader visual understanding capabilities by training on the M4-Instruct-data \cite{llavanext} and our \textit{AVR-Caption}.
    \item \textbf{Stage 3: General Reasoning and Active Visual Reasoning.} The final stage aims to instill general reasoning abilities and specialized active visual reasoning skills. For this, we fine-tune the model on a diverse mixture of datasets: Reason-RFT-129k \cite{Reasonrft}, AM-DeepSeek-R1-Distilled-100k \cite{AMr1Data}, our \textit{AVR-Embodied Reasoning} and \textit{AVR-Core} datasets.
\end{itemize}

This multi-stage training culminates in the \textbf{PhysVLM-AVR-3B} model. The detailed training configuration for PhysVLM-AVR-3B, including hyperparameters and software environment, is presented in Appendix Section~\ref{Appendix-model-train} and Figure~\ref{fig:training_details}.

\section{Experiment}

\subsection{Experimental Setup}

\textbf{Tasks.} We demonstrate the effectiveness of our dataset and model through experiments across the following three categories of tasks:
\begin{itemize}[leftmargin=2em]
    \item \textbf{Active Visual Reasoning.} To demonstrate that our approach enables models to acquire active visual reasoning capabilities, we first conduct comparative experiments on the \textit{CLEVR-AVR} (Sec.~\ref{sec:benchmark_dataset}) benchmark. The CLEVR-AVR benchmark is built on the Genesis simulation framework, ensuring that no similar images appear in the training data.
    \item \textbf{Embodied Reasoning and Planning.} To demonstrate that the reasoning capabilities developed with our dataset and exemplified by PhysVLM-AVR are also effective in embodied reasoning and task planning scenarios, we include comparative experiments on two embodied benchmarks: \textit{OpenEQA} \cite{openeqa} and \textit{RoboVQA} \cite{robovqa}.
    \item \textbf{Visual Reasoning.} To demonstrate the generalization reasoning capability of PhysVLM-AVR in static visual reasoning tasks, we further evaluate its performance on two visual structure perception benchmarks: \textit{GeoMath} \cite{mathllava} and \textit{Geometry3K} \cite{Geometry30k}.
\end{itemize}

\textbf{Baselines.} In addition to our \textbf{PhysVLM-AVR-3B} (Sec \ref{sec:physvlm_avr}), we also fine-tune Qwen2.5-VL-7B on the AVR-152K dataset to obtain \textbf{AVR-Qwen2.5-VL-7B} (fine-tune details see \ref{fig:training_details}). We compare them against four categories of models: (1) \texttt{Open-source MLLMs}: Qwen2.5-VL-7B \cite{qwen2vl} and LLaVA-OV-7B \cite{llavanext}, representing standard multimodal foundation models. (2)\texttt{Visual Reasoning MLLMs}: R1-Onevision-7B \cite{r1Onevision} and Reason-RFT-7B \cite{Reasonrft}, specialized for visual reasoning tasks. (3) \texttt{Embodied MLLMs}: Embodied-Reasoner-7B \cite{embodiedReasoner} and RoboBrain-7B \cite{robobrain}, designed for embodied reasoning tasks. (4) \texttt{API-based MLLMs}: GPT-4o \cite{gpt4} and Gemini-2.0-flash \cite{gemini}. 

\textbf{Evaluation Metrics.} For CLEVR-AVR, we report $ACC_{ISJ}$, $IGR$, and $ACC_{FA}$ (Sec \ref{Metrics}), measuring the correctness of judging whether the initial observation is sufficient, information gain rate of action decision and accuracy of final answers. For OpenEQA, we follow the \textit{LLM-score} \cite{openeqa} (GPT-4o) evaluation protocol of the original paper. For RoboVQA, we report \textit{BLEU1-4} scores \cite{robovqa} as established in the original benchmark.

\begin{table}[ht]
\caption{CLEVR-AVR Benchmark: Our PhysVLM-AVR-3B and AVR-Qwen2.5-VL-7B vs. various MLLM categories. Best bold, second underlined.}
\label{table1}
\begin{center}
\resizebox{\textwidth}{!}{
\begin{tabular}{l|cccccccccccc}
\toprule
& \multicolumn{3}{c}{Occlusion} & \multicolumn{3}{c}{Stack} & \multicolumn{3}{c}{Composite} & \multicolumn{3}{c}{AVG.} \\
Method  & $ACC_{ISJ}$ &  $IGR$ & $ACC_{FA}$  & $ACC_{ISJ}$ &  $IGR$ &$ACC_{FA}$  & $ACC_{ISJ}$  &  $IGR$ & $ACC_{FA}$ & $ACC_{ISJ}$  & $IGR$ & $ACC_{FA}$    \\ 
\midrule
\multicolumn{13}{l}{\texttt{Open-sourceMLLMs}} \\
LLaVA-OV-7B    & 0   & 0     & 0        & 0    & 0   & 0    & 0    & 0   & 0    & 0   & 0   & 0  \\
Qwen2.5-VL-7B  & 0   & 0     & 0        & 7.7  & 5.8 & 7.7  & 7.1  & 5.4 & 0    & 4.9 & 3.7 & 2.6 \\
\midrule
\multicolumn{13}{l}{\texttt{Reasoning MLLMs}} \\
R1-Onevision-7B & 0    & 0      & 0      & 6.6  & 4.9  & 3.3   & 6.1 & 6.1 &  2.0         & 4.2 & 3.7 & 1.8\\
Reason-RFT-7B   & 0    & 0      & 0      & 0    & 0    & 0     & 1.5 & 1.5 &  0.0         & 0.5 & 0.5 & 0.0   \\
\midrule
\multicolumn{13}{l}{\texttt{Embodied MLLMs}}\\
RoboBrain-7B            & 4.5   & 3.8 & 0.0         & 1.6   & 0.0 & 1.6     & 4.6 &  3.1  & 3.1     & 3.6  & 2.3 & 1.6  \\
Embodied-Reasoner-7B    & 21.2    & 13.6 & 1.5       & 16.4 & 8.2 & 3.3     & 23.1 & 10.8&  0.0     & 20.2  & 10.9 & 1.6   \\
\midrule
\multicolumn{13}{l}{\texttt{API-based MLLMs}}\\
Gemini-2.0-flash  & 50.8 & 24.4  & 33.6                     & 52.6 & \uline{27.4} & 31.0                        & 56.3 & \uline{42.0} & 25.9                    & 53.2 & 31.3 & 30.2 \\
GPT-4o            & 85.2  & \uline{39.4} & \textbf{53.1}    & \textbf{90.5} & \textbf{46.8} & \textbf{41.4}     & \uline{89.6} & \textbf{66.3} & \textbf{42.5}  & 88.4 & \textbf{50.8} & \textbf{45.7} \\
\midrule 
\rowcolor{blue!15}\textbf{AVR-Qwen2.5-VL-7B}  & \textbf{95.5} & \textbf{43.9}& 40.9      & \uline{88.5}  & 26.2  & 36.1             & 82.9 & 40.8 & 37.4                        & \uline{89.3} & \uline{34.7} & 38.1 \\
\rowcolor{blue!15}\textbf{PhysVLM-AVR-3B}  & \uline{90.6} & 27.4 & \uline{42.2}          & \textbf{90.5} & 22.4 & \uline{37.9}      & \textbf{90.2} & 40.0 & \uline{39.1}       & \textbf{90.5} & 29.9 & \uline{39.7} \\
\bottomrule
\end{tabular}
}
\end{center}
\vspace{-0.2cm}
\end{table}

\subsection{Results on Active Visual Reasoning Tasks}

Table \ref{table1} results from CLEVR-AVR offer compelling insights into active visual reasoning and highlight our AVR framework's significance. First, the results validate our premise: \textbf{existing MLLMs, trained on static data, struggle with active reasoning in interactive, partially observable environments.} Standard open-source MLLMs (LLaVA-OV-7B, Qwen2.5-VL-7B) and passive visual reasoning models (R1-Onevision-7B, Reason-RFT-7B) show near-zero performance. This illustrates passive capabilities don't translate to AVR's dynamic demands, necessitating new paradigms.

Critically, existing embodied MLLMs reveal a fundamental limitation. While models like Embodied-Reasoner-7B can detect information incompleteness (20.2\% $ACC_{ISJ}$), they largely fail to act effectively for correct reasoning (only 1.6\% $ACC_{FA}$). This highlights: \textbf{current embodied agents may recognize missing information but struggle to strategically acquire and integrate it.} 

In contrast, our PhysVLM-AVR-3B and the fine-tuned AVR-Qwen2.5-VL-7B excel in Information Sufficiency Judgment Accuracy ($ACC_{ISJ}$) (90.5\% and 89.3\%), surpassing GPT-4o (88.4\%). This \textbf{ validates the CoT of our AVR-Core dataset to teach the identification of uncertainty and the need for interaction.} Beyond this, AVR-Qwen2.5-VL-7B achieves a robust 39.7\% average final reasoning accuracy ($ACC_{FA}$), making it the best open source model, second to GPT-4o and well ahead of other baselines. The PhysVLM-AVR-3B also shows considerable improvement (39.7\% $ACC_{FA}$), further showing the impact of our data set.

However, the gap between our models' high $ACC_{ISJ}$ and their final $ACC_{FA}$ (e.g., PhysVLM-AVR-3B: 90.5\% $ACC_{ISJ}$ vs. 39.7\% $ACC_{FA}$) highlights AVR's central challenge: \textbf{mastering optimal action selection and multi-step information integration for coherent reasoning.} While identifying the \textit{need} to act is learned, consistently choosing the \textit{best} action and synthesizing information over time needs more development. 

In summary, CLEVR-AVR results show existing MLLMs' difficulty with AVR. They also affirm our AVR framework (AVR-152K dataset and PhysVLM-AVR model) is a significant step towards MLLMs that can intelligently explore, gather information, and reason in physical environments.

\subsection{Results on Embodied Reasoning and Planning Tasks}

\begin{figure*}[ht]
    \begin{center}
    \centerline{\includegraphics[width=1\columnwidth]{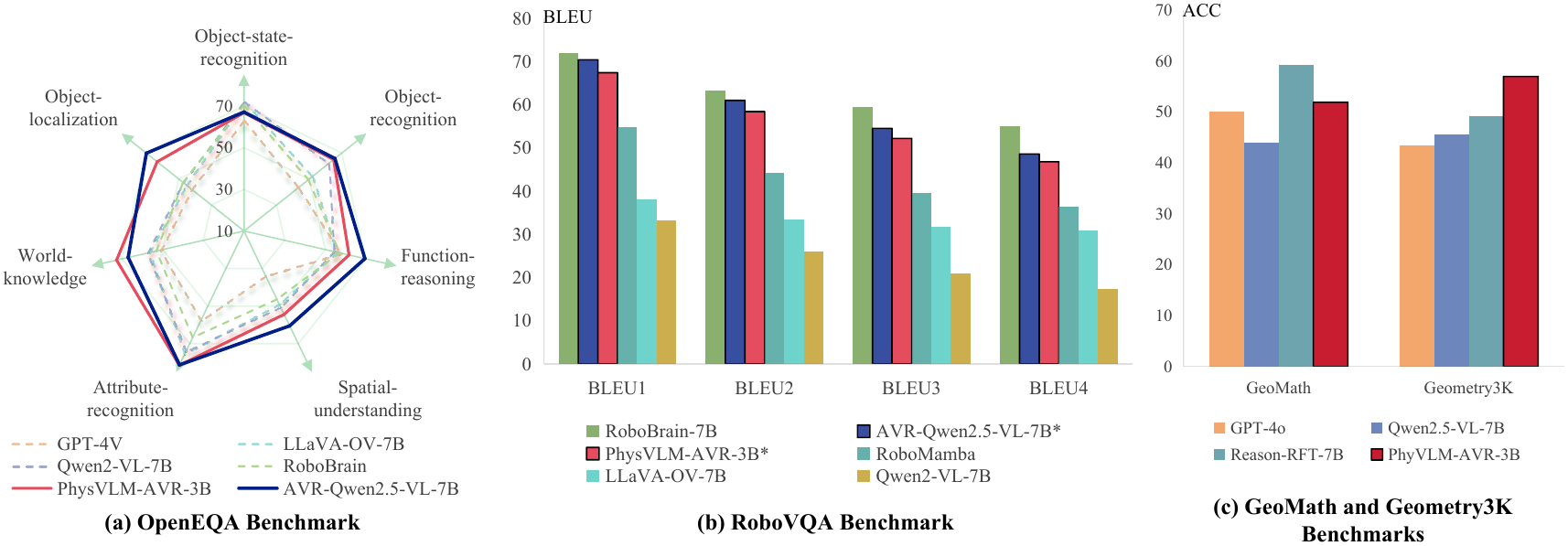}}
    \caption{Results for Embodied and Visual Reasoning Tasks.}
    \label{image4}
    \end{center}
    \vspace{-0.3cm}
\end{figure*}

As shown in Figure~\ref{image4}(a) and (b), our proposed models, PhysVLM-AVR-3B and AVR-Qwen2.5-VL-7B, achieve strong performance on both the OpenEQA and RoboVQA embodied reasoning benchmarks. On OpenEQA, our models consistently outperform standard multimodal models and dedicated embodied reasoning models across all sub-tasks. Notably, even when trained with only 1/20 of the RoboVQA training set (indicated by *), our models deliver BLEU scores close to or surpassing those of fully supervised baselines. These results demonstrate that active visual reasoning and the AVR-152k dataset significantly enhance embodied reasoning and planning capabilities, even under limited data conditions.

\subsection{Results on Visual Reasoning Tasks}

We further evaluate our approach on static visual reasoning benchmarks, GeoMath and Geometry3K. As shown in Figure~\ref{image4}(c), PhysVLM-AVR-3B achieves the highest Accuracy among all compared models, outperforming both large-scale API models (GPT-4o) and specialized visual reasoning models (Reason-RFT-7B). This indicates that our active reasoning paradigm not only benefits embodied scenarios but also generalizes well to traditional visual reasoning tasks, confirming the broad applicability and robustness of our approach.

\subsection{Ablation Study}

To assess the contributions of our AVR-Core dataset and its Chain-of-Thought (CoT) annotations, we conducted ablations on CLEVR-AVR (see Table \ref{table-ablation}). Removing the entire \textbf{AVR-Core} dataset ("w/o AVR-Core") caused a catastrophic performance collapse: $ACC_{ISJ}$ dropped from 90.5\% to 16.4\%, $ACC_{FA}$ from 39.1\% to 2.3\%, and $IGR$ from 29.9\% to 11.2\%. This highlights AVR-Core's fundamental role in teaching the model to actively gather information and reason iteratively within the higher-order MDP framework.

\begin{minipage}[t]{0.58\textwidth}
Excluding only the \textbf{Chain-of-Thought (CoT) annotations} from AVR-Core ("w/o CoT") also led to a significant decline: $ACC_{ISJ}$ fell to 47.6\%, $IGR$ to 18.0\%, and $ACC_{FA}$ to 16.9\%. This underscores the CoTs' crucial function in providing explicit supervision for the nuanced reasoning steps of uncertainty identification, action-conditioned information gain prediction, and strategic action selection.
\end{minipage}
\hfill
% \vspace{-3mm}
\begin{minipage}[t]{0.4\textwidth}
\vspace{0pt}
\centering
\footnotesize
% \begin{small}
\begin{tabular}{@{}lccc@{}}
\toprule
Method & $ACC_{ISJ}$  & $IGR$  & $ACC_{FA}$ \\
\midrule
Full Model    & 90.5 & 29.9 & 39.7 \\
w/o CoT       & 47.6 & 18.0 & 16.9 \\
w/o AVR-Core  & 16.4 & 11.2 & 2.3 \\
\bottomrule
\end{tabular}
% \end{small}
\captionof{table}{Ablation study results.}
\label{table-ablation}
\end{minipage}

These ablations clearly demonstrate that both the specialized AVR-Core dataset and its detailed CoT annotations are indispensable for developing effective Active Visual Reasoning capabilities.

\section{Conclusion}

In this work, we introduced Active Visual Reasoning (AVR), a novel paradigm that extends visual reasoning to interactive, partially observable environments. We developed the CLEVR-AVR benchmark to rigorously evaluate AVR capabilities and the AVR-152k dataset, with its core AVR-Core component providing detailed Chain-of-Thought annotations within a higher-order MDP framework, to train agents for this task. Our PhysVLM-AVR model demonstrates significant progress, achieving state-of-the-art performance on CLEVR-AVR and showing strong generalization to other embodied and static reasoning tasks. Our findings highlight that while current models can identify information incompleteness, a critical challenge remains in enabling them to strategically act to acquire and integrate new information effectively. Future work will focus on enhancing the model's ability to predict action-conditioned information gain and select optimal information-gathering actions. We also plan to explore the application of AVR to more complex real-world scenarios and investigate methods for improving sample efficiency in learning these active reasoning skills.

\section*{Acknowledgments}
This work was supported by the National Key Research and Development Program of China (No. 2022YFB4300400), the National Key R\&D Program of China under Grant Nos. 2024YFE0210600 and 2022ZD0160601, and the National Natural Science Foundation of China under Grant Nos. 62176254, 62276260.

\newpage

\bibliographystyle{unsrt}  %plainnat,abbrvnat,unsrtnat
\bibliography{Reference}

\begin{thebibliography}{10}

\bibitem{qwen2vl}
Peng Wang, Shuai Bai, Sinan Tan, Shijie Wang, Zhihao Fan, Jinze Bai, Keqin Chen, Xuejing Liu, Jialin Wang, Wenbin Ge, et~al.
\newblock Qwen2-vl: Enhancing vision-language model's perception of the world at any resolution.
\newblock {\em arXiv preprint arXiv:2409.12191}, 2024.

\bibitem{llava}
Haotian Liu, Chunyuan Li, Qingyang Wu, and Yong~Jae Lee.
\newblock Visual instruction tuning.
\newblock {\em Advances in neural information processing systems}, 36, 2024.

\bibitem{llavanext}
Feng Li, Renrui Zhang, Hao Zhang, Yuanhan Zhang, Bo~Li, Wei Li, Zejun Ma, and Chunyuan Li.
\newblock Llava-next-interleave: Tackling multi-image, video, and 3d in large multimodal models.
\newblock {\em arXiv preprint arXiv:2407.07895}, 2024.

\bibitem{internvl}
Zhe Chen, Jiannan Wu, Wenhai Wang, Weijie Su, Guo Chen, Sen Xing, Muyan Zhong, Qinglong Zhang, Xizhou Zhu, Lewei Lu, et~al.
\newblock Internvl: Scaling up vision foundation models and aligning for generic visual-linguistic tasks.
\newblock In {\em Proceedings of the IEEE/CVF Conference on Computer Vision and Pattern Recognition}, pages 24185--24198, 2024.

\bibitem{claude3}
Alec Radford, Jong~Wook Kim, Chris Hallacy, Aditya Ramesh, Gabriel Goh, Sandhini Agarwal, Girish Sastry, Amanda Askell, Pamela Mishkin, Jack Clark, et~al.
\newblock Learning transferable visual models from natural language supervision.
\newblock In {\em International conference on machine learning}, pages 8748--8763. PMLR, 2021.

\bibitem{CLEVR}
Justin Johnson, Bharath Hariharan, Laurens van~der Maaten, Li~Fei-Fei, C.~Lawrence~Zitnick, and Ross Girshick.
\newblock Clevr: A diagnostic dataset for compositional language and elementary visual reasoning.
\newblock In {\em Proceedings of the IEEE Conference on Computer Vision and Pattern Recognition (CVPR)}, July 2017.

\bibitem{cleve-math}
Ke~Wang, Junting Pan, Weikang Shi, Zimu Lu, Houxing Ren, Aojun Zhou, Mingjie Zhan, and Hongsheng Li.
\newblock Measuring multimodal mathematical reasoning with math-vision dataset.
\newblock {\em Advances in Neural Information Processing Systems}, 37:95095--95169, 2024.

\bibitem{SuperCLEVR}
Zhuowan Li, Xingrui Wang, Elias Stengel-Eskin, Adam Kortylewski, Wufei Ma, Benjamin Van~Durme, and Alan~L. Yuille.
\newblock Super-clevr: A virtual benchmark to diagnose domain robustness in visual reasoning.
\newblock In {\em Proceedings of the IEEE/CVF Conference on Computer Vision and Pattern Recognition (CVPR)}, pages 14963--14973, June 2023.

\bibitem{Visionr1}
Wenxuan Huang, Bohan Jia, Zijie Zhai, Shaosheng Cao, Zheyu Ye, Fei Zhao, Zhe Xu, Yao Hu, and Shaohui Lin.
\newblock Vision-r1: Incentivizing reasoning capability in multimodal large language models.
\newblock {\em arXiv preprint arXiv:2503.06749}, 2025.

\bibitem{VisionR1zhan}
Yufei Zhan, Yousong Zhu, Shurong Zheng, Hongyin Zhao, Fan Yang, Ming Tang, and Jinqiao Wang.
\newblock Vision-r1: Evolving human-free alignment in large vision-language models via vision-guided reinforcement learning, 2025.

\bibitem{Visualprogramming}
Tanmay Gupta and Aniruddha Kembhavi.
\newblock Visual programming: Compositional visual reasoning without training.
\newblock In {\em Proceedings of the IEEE/CVF Conference on Computer Vision and Pattern Recognition}, pages 14953--14962, 2023.

\bibitem{vqav2}
Yash Goyal, Tejas Khot, Douglas Summers{-}Stay, Dhruv Batra, and Devi Parikh.
\newblock Making the {V} in {VQA} matter: Elevating the role of image understanding in {V}isual {Q}uestion {A}nswering.
\newblock In {\em Conference on Computer Vision and Pattern Recognition (CVPR)}, 2017.

\bibitem{robobrain}
Yuheng Ji, Huajie Tan, Jiayu Shi, Xiaoshuai Hao, Yuan Zhang, Hengyuan Zhang, Pengwei Wang, Mengdi Zhao, Yao Mu, Pengju An, et~al.
\newblock Robobrain: A unified brain model for robotic manipulation from abstract to concrete.
\newblock {\em arXiv preprint arXiv:2502.21257}, 2025.

\bibitem{roboos}
Huajie Tan, Xiaoshuai Hao, Minglan Lin, Pengwei Wang, Yaoxu Lyu, Mingyu Cao, Zhongyuan Wang, and Shanghang Zhang.
\newblock Roboos: A hierarchical embodied framework for cross-embodiment and multi-agent collaboration, 2025.

\bibitem{physvlm}
Weijie Zhou, Manli Tao, Chaoyang Zhao, Haiyun Guo, Honghui Dong, Ming Tang, and Jinqiao Wang.
\newblock Physvlm: Enabling visual language models to understand robotic physical reachability, 2025.

\bibitem{physical1}
Yunsheng Tian, Karl~DD Willis, Bassel Al~Omari, Jieliang Luo, Pingchuan Ma, Yichen Li, Farhad Javid, Edward Gu, Joshua Jacob, Shinjiro Sueda, et~al.
\newblock Asap: Automated sequence planning for complex robotic assembly with physical feasibility.
\newblock In {\em 2024 IEEE International Conference on Robotics and Automation (ICRA)}, pages 4380--4386. IEEE, 2024.

\bibitem{okvqa}
Kenneth Marino, Mohammad Rastegari, Ali Farhadi, and Roozbeh Mottaghi.
\newblock Ok-vqa: A visual question answering benchmark requiring external knowledge.
\newblock In {\em Proceedings of the IEEE/cvf conference on computer vision and pattern recognition}, pages 3195--3204, 2019.

\bibitem{eqa1}
Yu~Wu, Lu~Jiang, and Yi~Yang.
\newblock Revisiting embodiedqa: A simple baseline and beyond.
\newblock {\em IEEE Transactions on Image Processing}, 29:3984--3992, 2020.

\bibitem{eqa2}
Yining Hong, Zishuo Zheng, Peihao Chen, Yian Wang, Junyan Li, and Chuang Gan.
\newblock Multiply: A multisensory object-centric embodied large language model in 3d world.
\newblock In {\em Proceedings of the IEEE/CVF Conference on Computer Vision and Pattern Recognition}, pages 26406--26416, 2024.

\bibitem{spatialvlm}
Boyuan Chen, Zhuo Xu, Sean Kirmani, Brain Ichter, Dorsa Sadigh, Leonidas Guibas, and Fei Xia.
\newblock Spatialvlm: Endowing vision-language models with spatial reasoning capabilities.
\newblock In {\em Proceedings of the IEEE/CVF Conference on Computer Vision and Pattern Recognition}, pages 14455--14465, 2024.

\bibitem{spatialbot}
Wenxiao Cai, Yaroslav Ponomarenko, Jianhao Yuan, Xiaoqi Li, Wankou Yang, Hao Dong, and Bo~Zhao.
\newblock Spatialbot: Precise spatial understanding with vision language models.
\newblock {\em arXiv preprint arXiv:2406.13642}, 2024.

\bibitem{geminirobotics}
Gemini~Robotics Team, Saminda Abeyruwan, et~al.
\newblock Gemini robotics: Bringing ai into the physical world, 2025.

\bibitem{physical2}
Giovanni Sutanto, Austin Wang, Yixin Lin, Mustafa Mukadam, Gaurav Sukhatme, Akshara Rai, and Franziska Meier.
\newblock Encoding physical constraints in differentiable newton-euler algorithm.
\newblock In Alexandre~M. Bayen, Ali Jadbabaie, George Pappas, Pablo~A. Parrilo, Benjamin Recht, Claire Tomlin, and Melanie Zeilinger, editors, {\em Proceedings of the 2nd Conference on Learning for Dynamics and Control}, volume 120 of {\em Proceedings of Machine Learning Research}, pages 804--813. PMLR, 10--11 Jun 2020.

\bibitem{Reasonrft}
Huajie Tan, Yuheng Ji, Xiaoshuai Hao, Minglan Lin, Pengwei Wang, Zhongyuan Wang, and Shanghang Zhang.
\newblock Reason-rft: Reinforcement fine-tuning for visual reasoning.
\newblock {\em arXiv preprint arXiv:2503.20752}, 2025.

\bibitem{mmbench}
Yuan Liu, Haodong Duan, Yuanhan Zhang, Bo~Li, Songyang Zhang, Wangbo Zhao, Yike Yuan, Jiaqi Wang, Conghui He, Ziwei Liu, et~al.
\newblock Mmbench: Is your multi-modal model an all-around player?
\newblock In {\em European Conference on Computer Vision}, pages 216--233. Springer, 2025.

\bibitem{mmvet}
Weihao Yu, Zhengyuan Yang, Linjie Li, Jianfeng Wang, Kevin Lin, Zicheng Liu, Xinchao Wang, and Lijuan Wang.
\newblock Mm-vet: Evaluating large multimodal models for integrated capabilities.
\newblock {\em arXiv preprint arXiv:2308.02490}, 2023.

\bibitem{mmebenchmark}
Chaoyou Fu, Peixian Chen, Yunhang Shen, Yulei Qin, Mengdan Zhang, Xu~Lin, Jinrui Yang, Xiawu Zheng, Ke~Li, Xing Sun, Yunsheng Wu, and Rongrong Ji.
\newblock Mme: A comprehensive evaluation benchmark for multimodal large language models, 2024.

\bibitem{openeqa}
Arjun Majumdar, Anurag Ajay, Xiaohan Zhang, Pranav Putta, Sriram Yenamandra, Mikael Henaff, Sneha Silwal, Paul Mcvay, Oleksandr Maksymets, Sergio Arnaud, et~al.
\newblock Openeqa: Embodied question answering in the era of foundation models.
\newblock In {\em Proceedings of the IEEE/CVF Conference on Computer Vision and Pattern Recognition}, pages 16488--16498, 2024.

\bibitem{robovqa}
Pierre Sermanet, Tianli Ding, Jeffrey Zhao, Fei Xia, Debidatta Dwibedi, Keerthana Gopalakrishnan, Christine Chan, Gabriel Dulac-Arnold, Sharath Maddineni, Nikhil~J Joshi, et~al.
\newblock Robovqa: Multimodal long-horizon reasoning for robotics.
\newblock In {\em 2024 IEEE International Conference on Robotics and Automation (ICRA)}, pages 645--652. IEEE, 2024.

\bibitem{askforhelp}
Ram Ramrakhya, Matthew Chang, Xavier Puig, Ruta Desai, Zsolt Kira, and Roozbeh Mottaghi.
\newblock Grounding multimodal llms to embodied agents that ask for help with reinforcement learning.
\newblock {\em arXiv preprint arXiv:2504.00907}, 2025.

\bibitem{embodiedReasoner}
Wenqi Zhang, Mengna Wang, Gangao Liu, Xu~Huixin, Yiwei Jiang, Yongliang Shen, Guiyang Hou, Zhe Zheng, Hang Zhang, Xin Li, et~al.
\newblock Embodied-reasoner: Synergizing visual search, reasoning, and action for embodied interactive tasks.
\newblock {\em arXiv preprint arXiv:2503.21696}, 2025.

\bibitem{Knowledge-based-eqa}
Sinan Tan, Mengmeng Ge, Di~Guo, Huaping Liu, and Fuchun Sun.
\newblock Knowledge-based embodied question answering.
\newblock {\em IEEE Transactions on Pattern Analysis and Machine Intelligence}, 45(10):11948--11960, 2023.

\bibitem{insightv}
Yuhao Dong, Zuyan Liu, Hai-Long Sun, Jingkang Yang, Winston Hu, Yongming Rao, and Ziwei Liu.
\newblock Insight-v: Exploring long-chain visual reasoning with multimodal large language models, 2025.

\bibitem{awais2024mathvision}
Muhammad Awais, Tauqir Ahmed, Muhammad Aslam, Amjad Rehman, Faten~S Alamri, Saeed~Ali Bahaj, and Tanzila Saba.
\newblock Mathvision: An accessible intelligent agent for visually impaired people to understand mathematical equations.
\newblock {\em IEEE Access}, 2024.

\bibitem{robospatial}
Chan~Hee Song, Valts Blukis, Jonathan Tremblay, Stephen Tyree, Yu~Su, and Stan Birchfield.
\newblock Robospatial: Teaching spatial understanding to 2d and 3d vision-language models for robotics.
\newblock {\em arXiv preprint arXiv:2411.16537}, 2024.

\bibitem{cityeqa}
Yong Zhao, Kai Xu, Zhengqiu Zhu, Yue Hu, Zhiheng Zheng, Yingfeng Chen, Yatai Ji, Chen Gao, Yong Li, and Jincai Huang.
\newblock Cityeqa: A hierarchical llm agent on embodied question answering benchmark in city space.
\newblock {\em arXiv preprint arXiv:2502.12532}, 2025.

\bibitem{etPlanBench}
Lingfeng Zhang, Yuening Wang, Hongjian Gu, Atia Hamidizadeh, Zhanguang Zhang, Yuecheng Liu, Yutong Wang, David Gamaliel~Arcos Bravo, Junyi Dong, Shunbo Zhou, et~al.
\newblock Et-plan-bench: Embodied task-level planning benchmark towards spatial-temporal cognition with foundation models.
\newblock {\em arXiv preprint arXiv:2410.14682}, 2024.

\bibitem{expressBench}
Kaixuan Jiang, Yang Liu, Weixing Chen, Jingzhou Luo, Ziliang Chen, Ling Pan, Guanbin Li, and Liang Lin.
\newblock Beyond the destination: A novel benchmark for exploration-aware embodied question answering.
\newblock {\em arXiv preprint arXiv:2503.11117}, 2025.

\bibitem{Genesis}
Genesis Authors.
\newblock Genesis: A universal and generative physics engine for robotics and beyond, December 2024.

\bibitem{scannet}
Angela Dai, Angel~X Chang, Manolis Savva, Maciej Halber, Thomas Funkhouser, and Matthias Nie{\ss}ner.
\newblock Scannet: Richly-annotated 3d reconstructions of indoor scenes.
\newblock In {\em Proceedings of the IEEE conference on computer vision and pattern recognition}, pages 5828--5839, 2017.

\bibitem{rt1}
Anthony Brohan, Noah Brown, et~al.
\newblock Rt-1: Robotics transformer for real-world control at scale, 2023.

\bibitem{gemini}
Gemini Team, Petko Georgiev, Ving~Ian Lei, Ryan Burnell, Libin Bai, Anmol Gulati, Garrett Tanzer, Damien Vincent, Zhufeng Pan, Shibo Wang, et~al.
\newblock Gemini 1.5: Unlocking multimodal understanding across millions of tokens of context.
\newblock {\em arXiv preprint arXiv:2403.05530}, 2024.

\bibitem{deepseekr1}
Daya Guo, Dejian Yang, Haowei Zhang, Junxiao Song, Ruoyu Zhang, Runxin Xu, Qihao Zhu, Shirong Ma, Peiyi Wang, Xiao Bi, et~al.
\newblock Deepseek-r1: Incentivizing reasoning capability in llms via reinforcement learning.
\newblock {\em arXiv preprint arXiv:2501.12948}, 2025.

\bibitem{deepseekv3}
Aixin Liu, Bei Feng, Bing Xue, Bingxuan Wang, Bochao Wu, Chengda Lu, Chenggang Zhao, Chengqi Deng, Chenyu Zhang, Chong Ruan, et~al.
\newblock Deepseek-v3 technical report.
\newblock {\em arXiv preprint arXiv:2412.19437}, 2024.

\bibitem{umi}
Cheng Chi, Zhenjia Xu, Chuer Pan, Eric Cousineau, Benjamin Burchfiel, Siyuan Feng, Russ Tedrake, and Shuran Song.
\newblock Universal manipulation interface: In-the-wild robot teaching without in-the-wild robots.
\newblock {\em arXiv preprint arXiv:2402.10329}, 2024.

\bibitem{qwen2.5}
Qwen Team.
\newblock qwen2.5, September 2024.

\bibitem{siglip}
Xiaohua Zhai, Basil Mustafa, Alexander Kolesnikov, and Lucas Beyer.
\newblock Sigmoid loss for language image pre-training, 2023.

\bibitem{AMr1Data}
Han Zhao, Haotian Wang, Yiping Peng, Sitong Zhao, Xiaoyu Tian, Shuaiting Chen, Yunjie Ji, and Xiangang Li.
\newblock 1.4 million open-source distilled reasoning dataset to empower large language model training, 2025.

\bibitem{mathllava}
Wenhao Shi, Zhiqiang Hu, Yi~Bin, Junhua Liu, Yang Yang, See-Kiong Ng, Lidong Bing, and Roy Ka-Wei Lee.
\newblock Math-llava: Bootstrapping mathematical reasoning for multimodal large language models, 2024.

\bibitem{Geometry30k}
Pan Lu, Ran Gong, Shibiao Jiang, Liang Qiu, Siyuan Huang, Xiaodan Liang, and Song-Chun Zhu.
\newblock Inter-gps: Interpretable geometry problem solving with formal language and symbolic reasoning.
\newblock In {\em The Joint Conference of the 59th Annual Meeting of the Association for Computational Linguistics and the 11th International Joint Conference on Natural Language Processing (ACL-IJCNLP 2021)}, 2021.

\bibitem{r1Onevision}
Yi~Yang, Xiaoxuan He, Hongkun Pan, Xiyan Jiang, Yan Deng, Xingtao Yang, Haoyu Lu, Dacheng Yin, Fengyun Rao, Minfeng Zhu, et~al.
\newblock R1-onevision: Advancing generalized multimodal reasoning through cross-modal formalization.
\newblock {\em arXiv preprint arXiv:2503.10615}, 2025.

\bibitem{gpt4}
Josh Achiam, Steven Adler, Sandhini Agarwal, Lama Ahmad, Ilge Akkaya, Florencia~Leoni Aleman, Diogo Almeida, Janko Altenschmidt, Sam Altman, Shyamal Anadkat, et~al.
\newblock Gpt-4 technical report.
\newblock {\em arXiv preprint arXiv:2303.08774}, 2023.

\end{thebibliography}
\normalsize

%%%%%%%%%%%%%%%%%%%%%%%%%%%%%%%%%%%%%%%%%%%%%%%%%%%%%%%%%%%%
\newpage
\clearpage

\appendix

\renewcommand{\thefigure}{A-\arabic{figure}}
\setcounter{figure}{0} % Reset figure counter for appendix

\section{More details of CLEVR-AVR benchmark}

\begin{figure*}[ht]
    \begin{center}
    \centerline{\includegraphics[width=1\columnwidth]{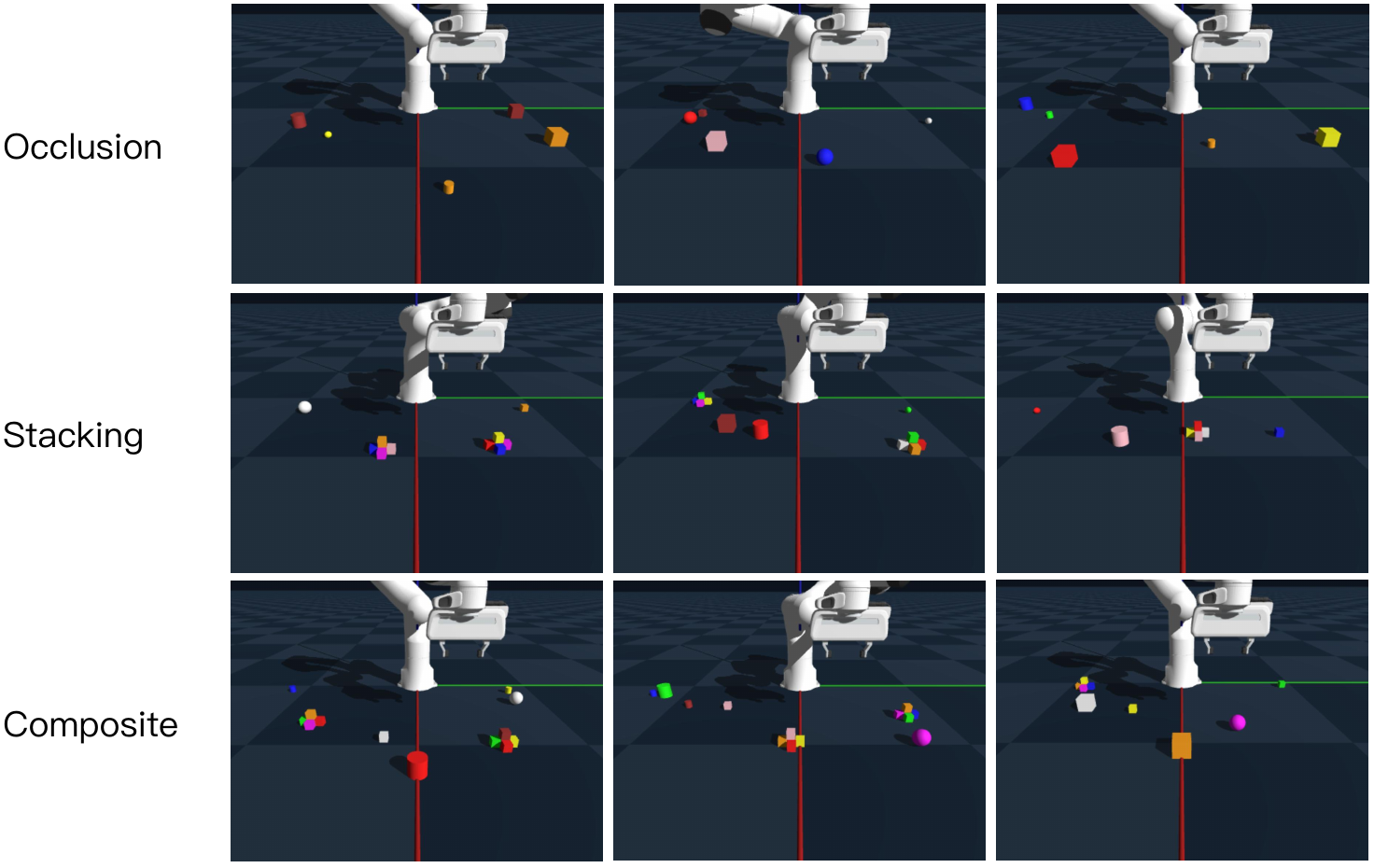}}
    \caption{Examples of the three scene types in the CLEVR-AVR benchmark.}
    \label{fig:scene_examples}
    \end{center}
\end{figure*}

\begin{figure*}[ht]
    \begin{center}
    \centerline{\includegraphics[width=1\columnwidth]{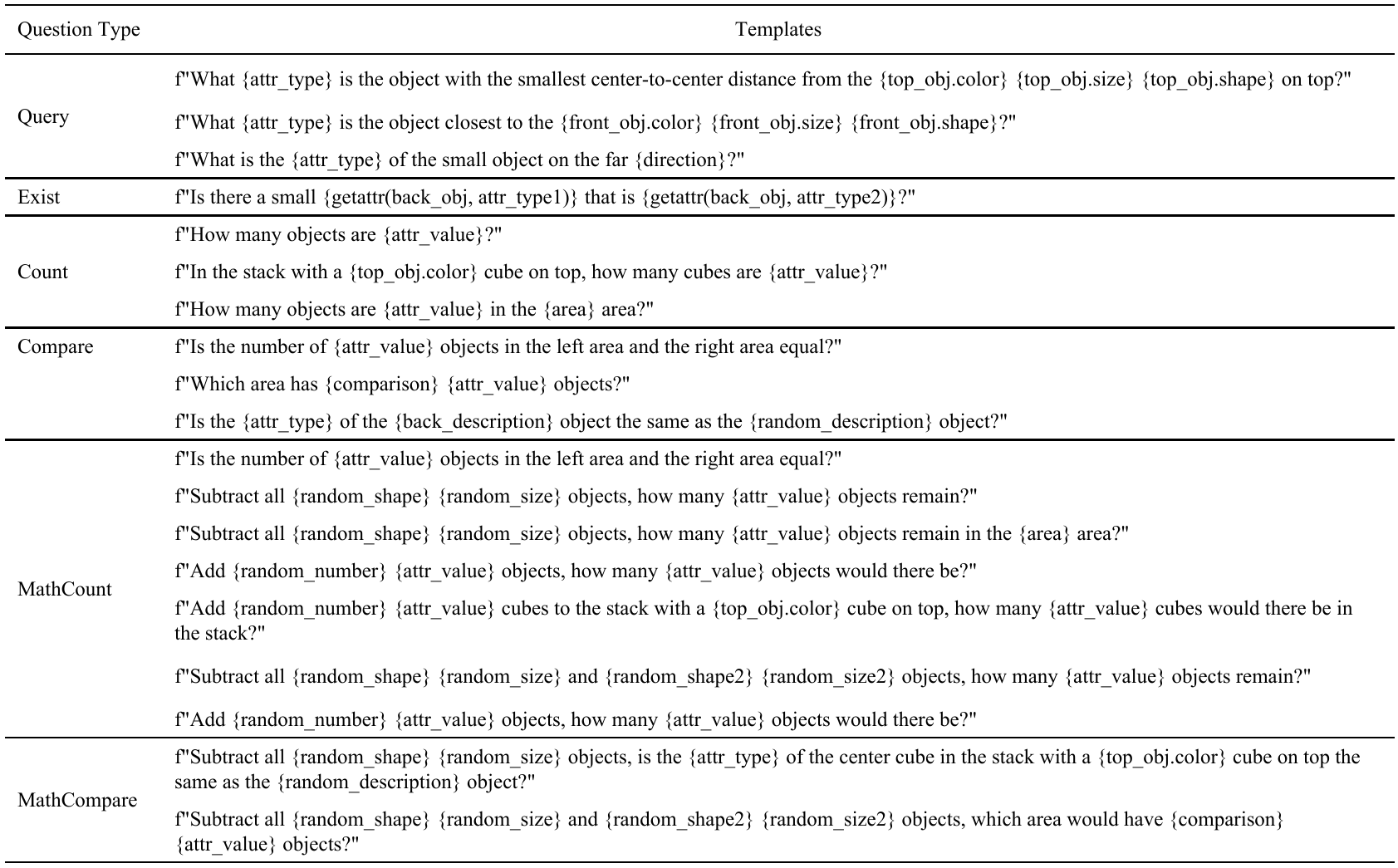}}
    \caption{Question template settings for different question types in the CLEVR-AVR benchmark.}
    \label{fig:question_templates}
    \end{center}
\end{figure*}

\begin{figure*}[ht]
    \begin{center}
    \centerline{\includegraphics[width=1\columnwidth]{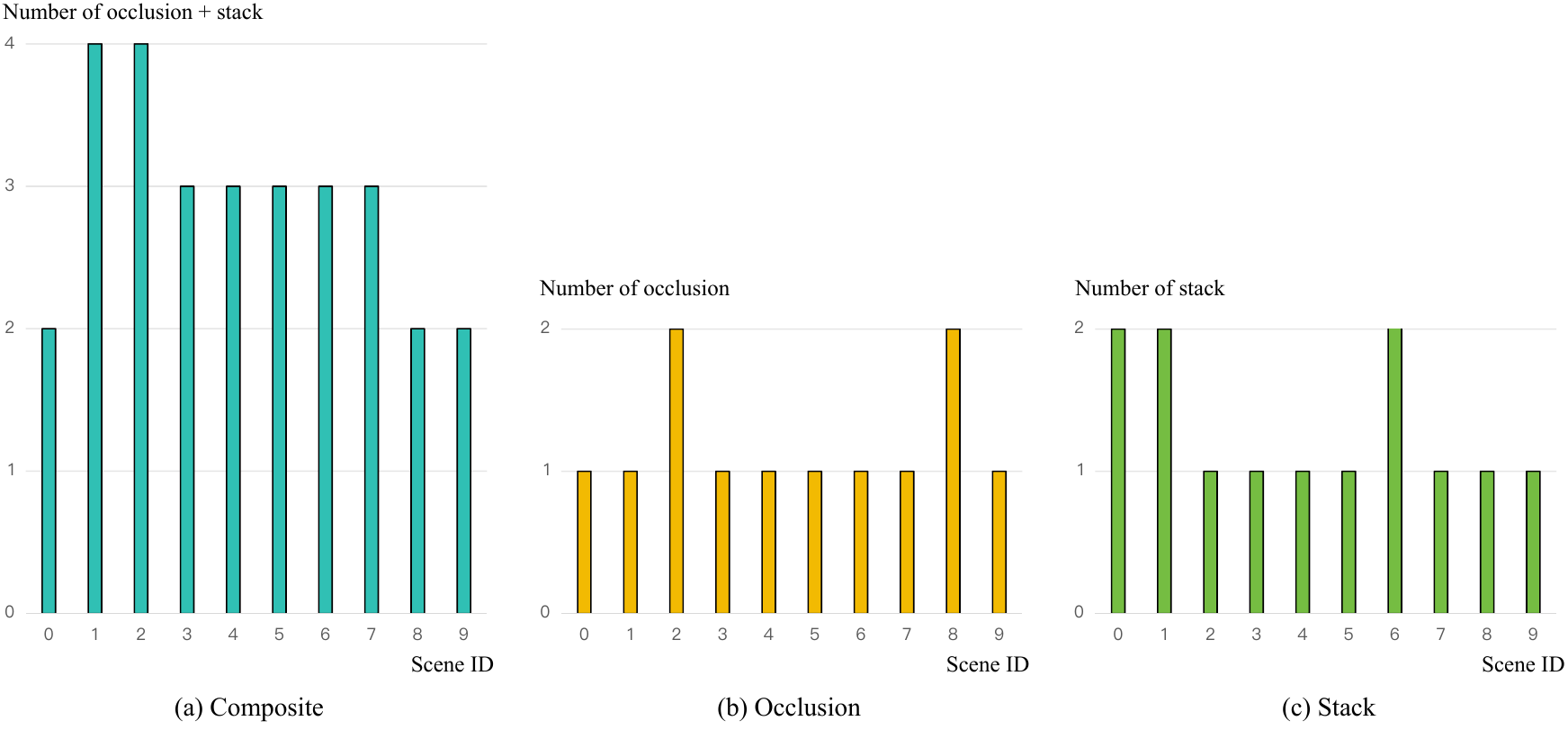}}
    \caption{Distribution of occlusion and stacking quantities across the three scene types in the CLEVR-AVR benchmark.}
    \label{fig:occlusion_distribution}
    \end{center}
\end{figure*}

Figure~\ref{fig:scene_examples} shows examples of the three scene types in the CLEVR-AVR benchmark, Figure~\ref{fig:question_templates} displays the question template settings for different question types, and Figure~\ref{fig:occlusion_distribution} illustrates the distribution of occlusion and stacking quantities in each of the three scene types.

\subsection{Action Candidate Generation in CLEVR-AVR}
\label{appendix:action_generation_details}

The available action space includes Pick, Move Viewer, Rotate Viewer, and Move Object. Beyond the diverse scenarios, the benchmark incorporates a rich variety of question types, including Query, Exist, Counting, Compare, Math Counting, and Math Compare. Agents must utilize the provided Franka robotic manipulator and camera controls to actively uncover essential details, thereby tackling the challenge of efficient exploration under conditions of partial observability. The agent communicates its decision by generating text that includes its reasoning (CoT) and the selected action, formatted for example as <action>E</action>, where E would map to a specific action like Pick(yellow cube) from the candidate list.

At each interaction step in CLEVR-AVR, the agent is presented with 5-8 candidate actions. The types of actions (e.g., Pick, Move Viewer, Rotate Viewer, Move Object) are predefined. However, the target objects for actions like Pick or Move Object are dynamically selected based on the current visual scene, often including objects relevant to resolving potential occlusions or ambiguities. To rigorously test the agent's reasoning-driven action selection, approximately 3-5 of these candidate actions are intentionally designed as distractors or sub-optimal choices that would yield less information gain towards answering the question. This forces the model to not just pick any valid action, but to strategically select the one most likely to resolve its current uncertainty.

\section{More details of AVR-152k dataset}

\begin{figure*}[ht]
    \begin{center}
    \centerline{\includegraphics[width=1\columnwidth]{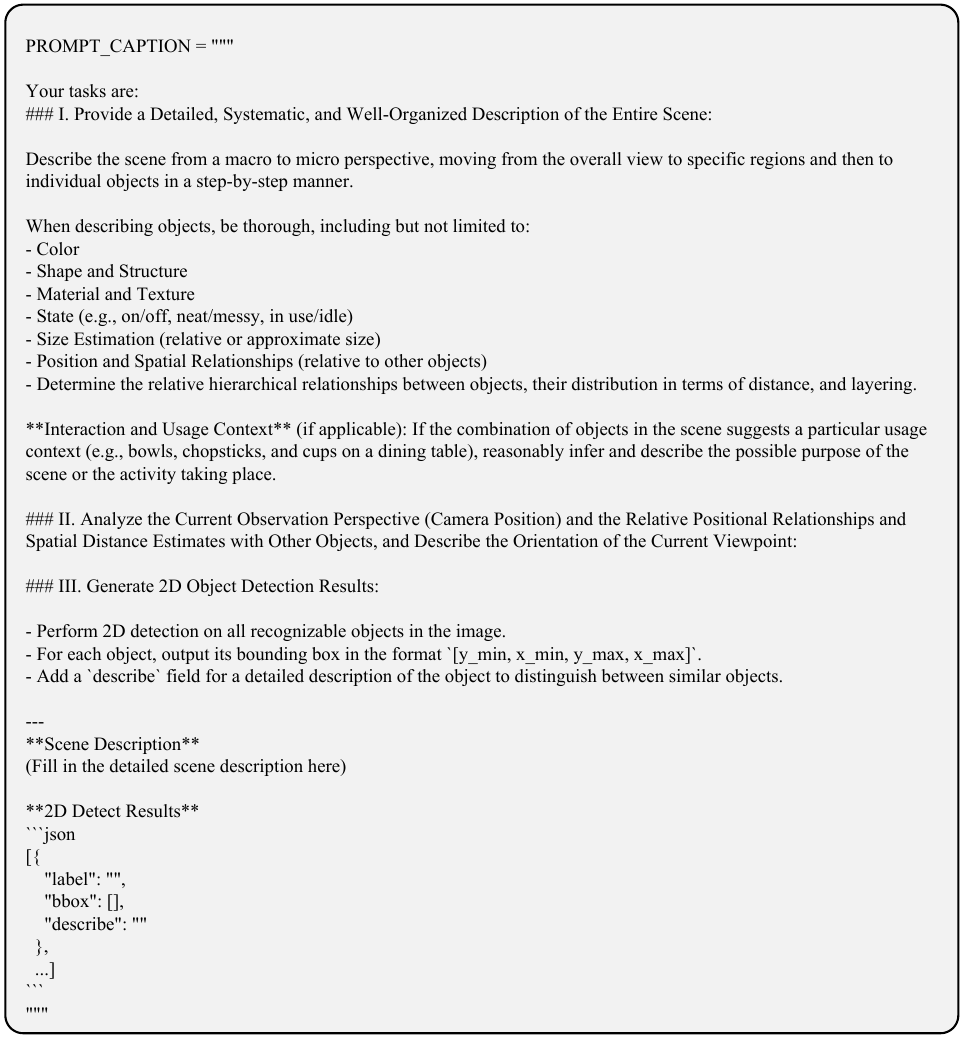}}
    \caption{System prompt for generating AVR-Caption data.}
    \label{fig:caption_prompt}
    \end{center}
\end{figure*}

\begin{figure*}[ht]
    \begin{center}
    \centerline{\includegraphics[width=1\columnwidth]{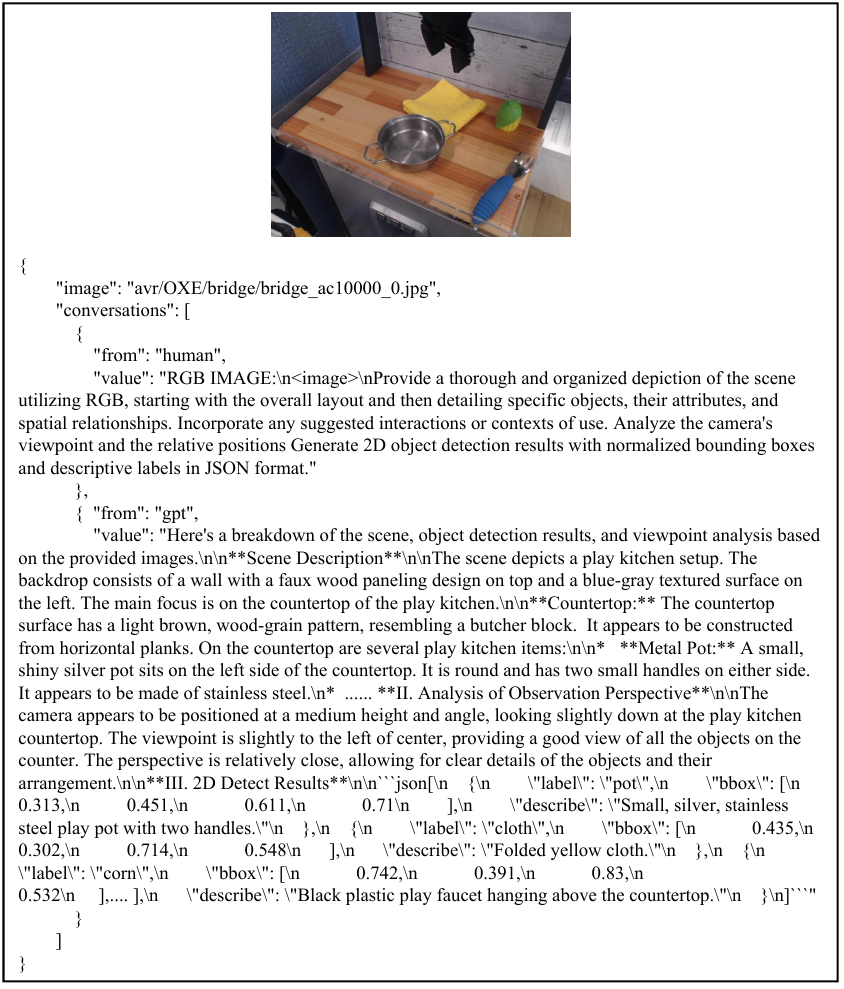}}
    \caption{Data instance of AVR-Caption.}
    \label{fig:data-caption}
    \end{center}
\end{figure*}

Figure~\ref{fig:caption_prompt} shows the system prompt used for generating AVR-Caption data. An example of an AVR-Caption data instance, featuring an image and its corresponding dense caption, is presented in Figure~\ref{fig:data-caption}.

\begin{figure*}[ht]
    \begin{center}
    \centerline{\includegraphics[width=1\columnwidth]{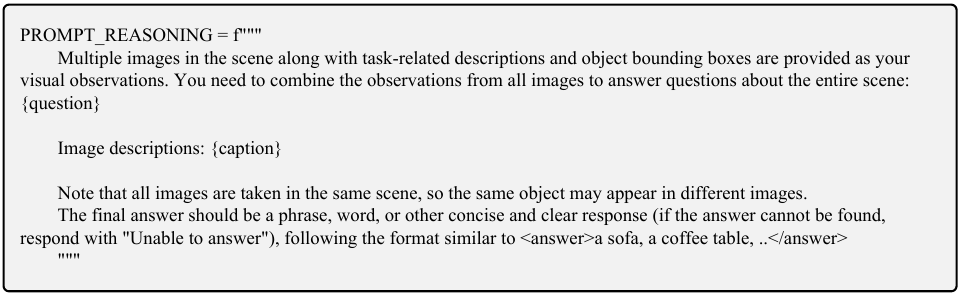}}
    \caption{Prompt for DeepSeek-R1 reasoning in AVR-Embodied Reasoning data.}
    \label{fig:embodied_reasoning_prompt}
    \end{center}
\end{figure*}

\begin{figure*}[ht]
    \begin{center}
    \centerline{\includegraphics[width=1\columnwidth]{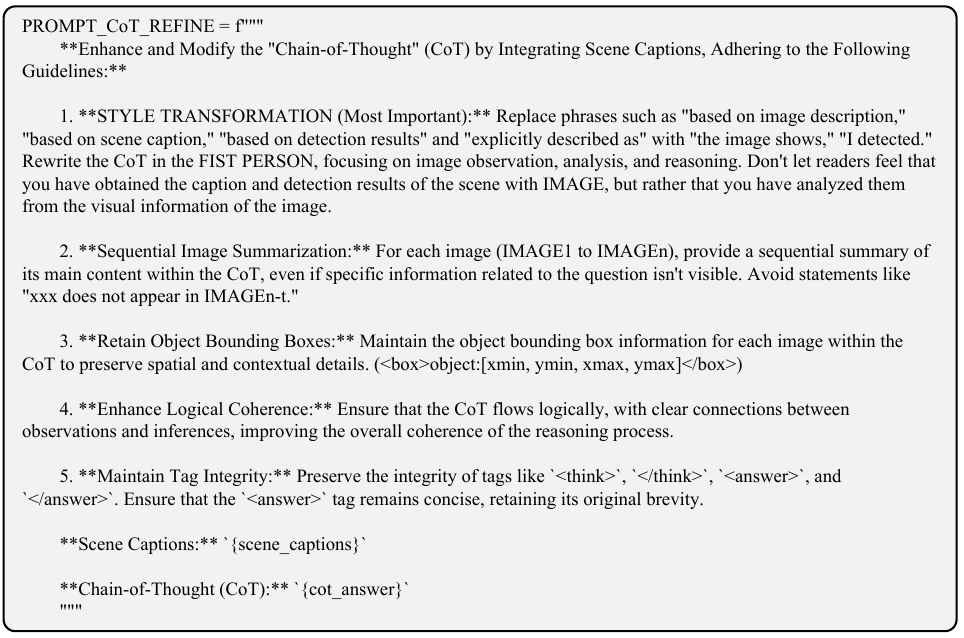}}
    \caption{Prompt for DeepSeek-V3 to refine Chain-of-Thought (CoT) reasoning in AVR-Embodied Reasoning data.}
    \label{fig:cot_refine_prompt}
    \end{center}
\end{figure*}

\begin{figure*}[ht]
    \begin{center}
    \centerline{\includegraphics[width=1\columnwidth]{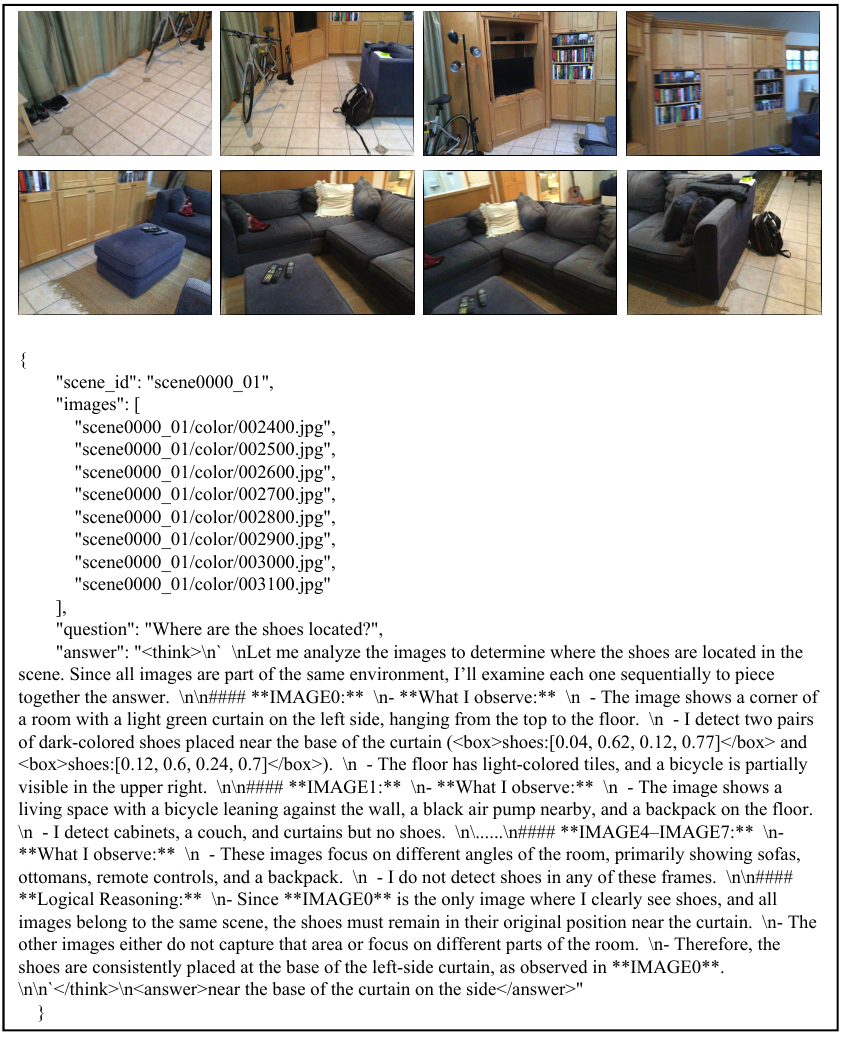}}
    \caption{Data instance of AVR-Embodied Reasoning.}
    \label{fig:data-embodied-reasoning}
    \end{center}
\end{figure*}

For the AVR-Embodied Reasoning subset, Figure~\ref{fig:embodied_reasoning_prompt} displays the prompt used for DeepSeek-R1 to generate initial reasoning, and Figure~\ref{fig:cot_refine_prompt} illustrates the prompt for DeepSeek-V3 to refine this Chain-of-Thought (CoT) reasoning. A representative data instance from AVR-Embodied Reasoning, which includes a multi-image sequence, a question, and the associated reasoning chain, can be seen in Figure~\ref{fig:data-embodied-reasoning}.

\begin{figure*}[ht]
    \begin{center}
    \centerline{\includegraphics[width=1\columnwidth]{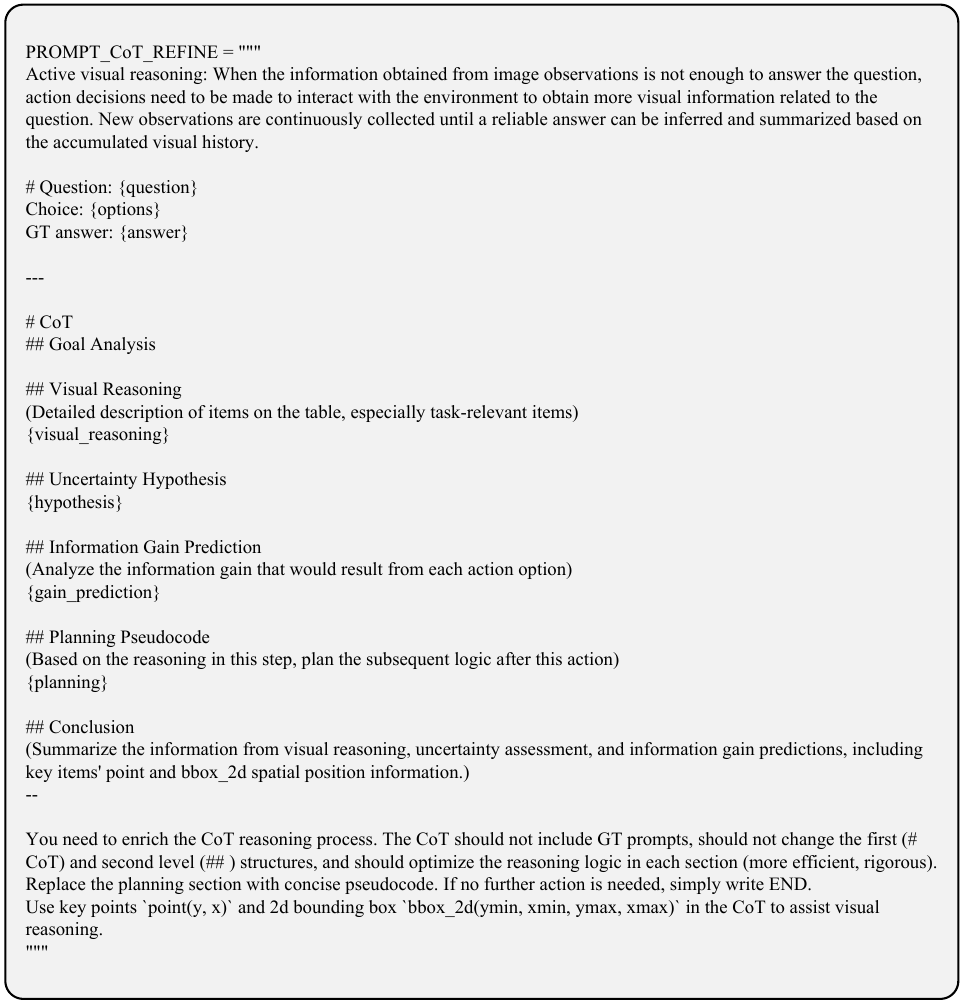}}
    \caption{Prompt for Gemini to refine Chain-of-Thought annotations in AVR-Core data, where \{question\}, \{options\}, \{answer\}, \{visual\_reasoning\}, \{hypothesis\}, \{gain\_prediction\}, and \{planning\} are all derived from expert human annotations.}
    \label{fig:gemini_refine_prompt}
    \end{center}
\end{figure*}

The prompt provided to Gemini for refining human expert Chain-of-Thought annotations within the AVR-Core dataset is detailed in Figure~\ref{fig:gemini_refine_prompt}. This prompt utilizes placeholders such as \{question\}, \{options\}, \{answer\}, \{visual\_reasoning\}, \{hypothesis\}, \{gain\_prediction\}, and \{planning\}, which are filled with content from expert human annotations. To concretely illustrate the rich, multi-step interactive reasoning process captured in AVR-Core, Figures~\ref{fig:data-core-step0}, \ref{fig:data-core-step1}, and \ref{fig:data-core-step2} collectively showcase a complete, sequential example from an AVR-Core instance.

\subsection{Initial Generation of Expert CoT Annotations for AVR-Core}
\label{appendix:cot_generation_detail}

The Chain-of-Thought (CoT) annotations in the AVR-Core dataset, prior to their refinement by large language models like Gemini, were meticulously crafted by human experts. The process involved two main stages:

\textbf{Live Task Execution and Initial Logging:} Human experts actively performed the interactive tasks using the UMI devices in real-world tabletop settings. During this live interaction, they made contemporaneous notes capturing their key reasoning steps, uncertainties identified, hypotheses about hidden information, and the rationale behind their intended actions at each decision point.

\textbf{Post-Interaction Refinement and Structuring:} After completing each task, the experts revisited their initial logs alongside the complete record of the interaction (including all visual observations and executed actions). They then elaborated on their initial notes, structuring them into the detailed, step-by-step CoT format that reflects the iterative process of assessing information, predicting gains from potential actions, and making a decision.

This human-centric initial annotation phase was crucial for ensuring that the CoTs genuinely reflected plausible and effective human reasoning strategies for active information gathering, forming a high-quality foundation for subsequent automated refinement and model training.

\begin{figure*}[ht]
    \begin{center}
    \centerline{\includegraphics[width=1\columnwidth]{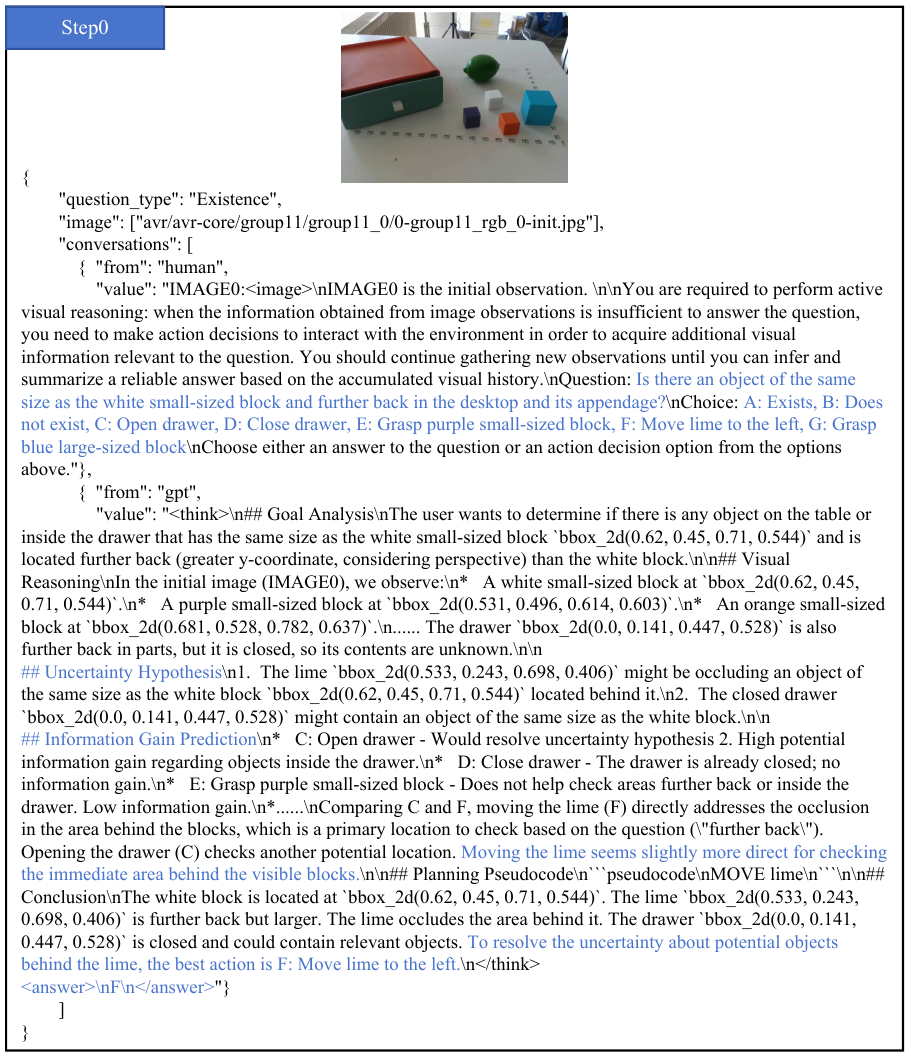}}
    \caption{Data instance of AVR-Embodied Reasoning. Step 0 of the active visual reasoning process, which make a action decision to get more information.}
    \label{fig:data-core-step0}
    \end{center}
\end{figure*}

\begin{figure*}[ht]
    \begin{center}
    \centerline{\includegraphics[width=1\columnwidth]{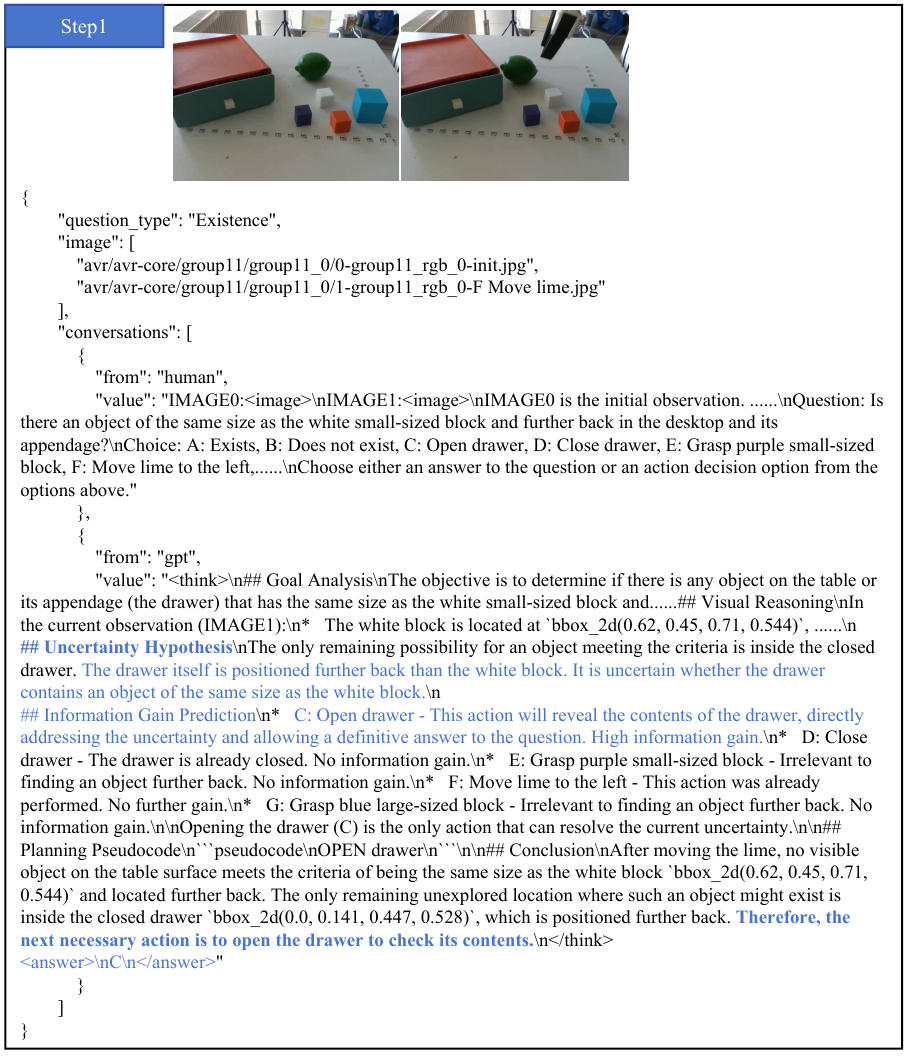}}
    \caption{Data instance of AVR-Embodied Reasoning. Step 1 of the active visual reasoning process, which make a action decision to get more information.}
    \label{fig:data-core-step1}
    \end{center}
\end{figure*}

\begin{figure*}[ht]
    \begin{center}
    \centerline{\includegraphics[width=1\columnwidth]{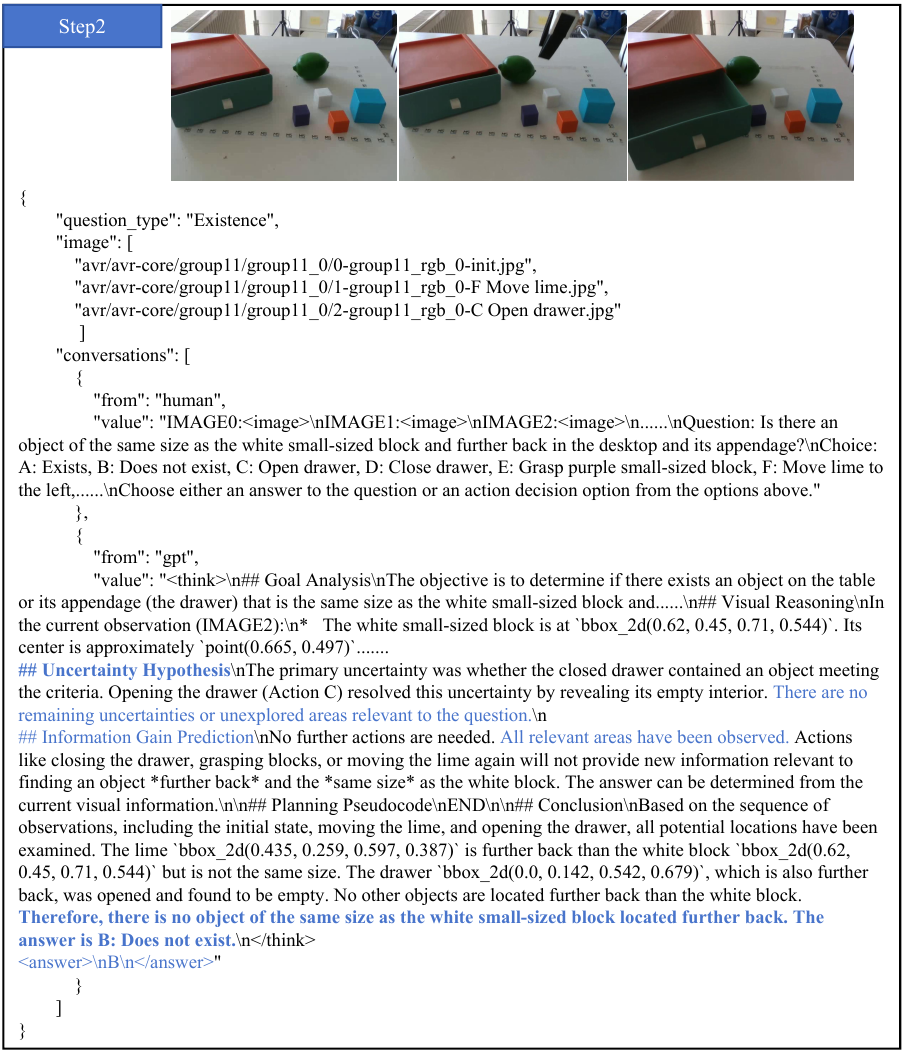}}
    \caption{Data instance of AVR-Embodied Reasoning. Step 2 of the active visual reasoning process, which getting the final answer.}
    \label{fig:data-core-step2}
    \end{center}
\end{figure*}

\begin{itemize}
    \item Specifically, Figure~\ref{fig:data-core-step0} depicts the initial state (Step 0) of an active visual reasoning scenario. Here, based on the initial visual observation and the posed question, the agent's CoT reflects its assessment of information insufficiency and its decision to take an action ("Move the yellow object to the left") to acquire more details.
    \item Figure~\ref{fig:data-core-step1} shows the subsequent state (Step 1) after the execution of the first action. The agent re-evaluates the situation with the new visual input (IMAGE1), and its CoT again indicates the need for further information to fully resolve the question, leading to the planning of another action ("Move the green object to the right").
    \item Finally, Figure~\ref{fig:data-core-step2} illustrates Step 2 of the process. After the second action and observing IMAGE2, the agent has gathered sufficient visual evidence, and its CoT culminates in providing the final answer ("Yes") to the question.
\end{itemize}

Together, these three figures (Figures~\ref{fig:data-core-step0}-\ref{fig:data-core-step2}) highlight the core principles of AVR-Core: the iterative cycle of identifying uncertainty, predicting information gain conditioned on potential actions, and making strategic decisions, all articulated through detailed CoT annotations.

\section{More details of Model and Training}
\label{Appendix-model-train}

\begin{figure*}[ht]
    \begin{center}
    \centerline{\includegraphics[width=1\columnwidth]{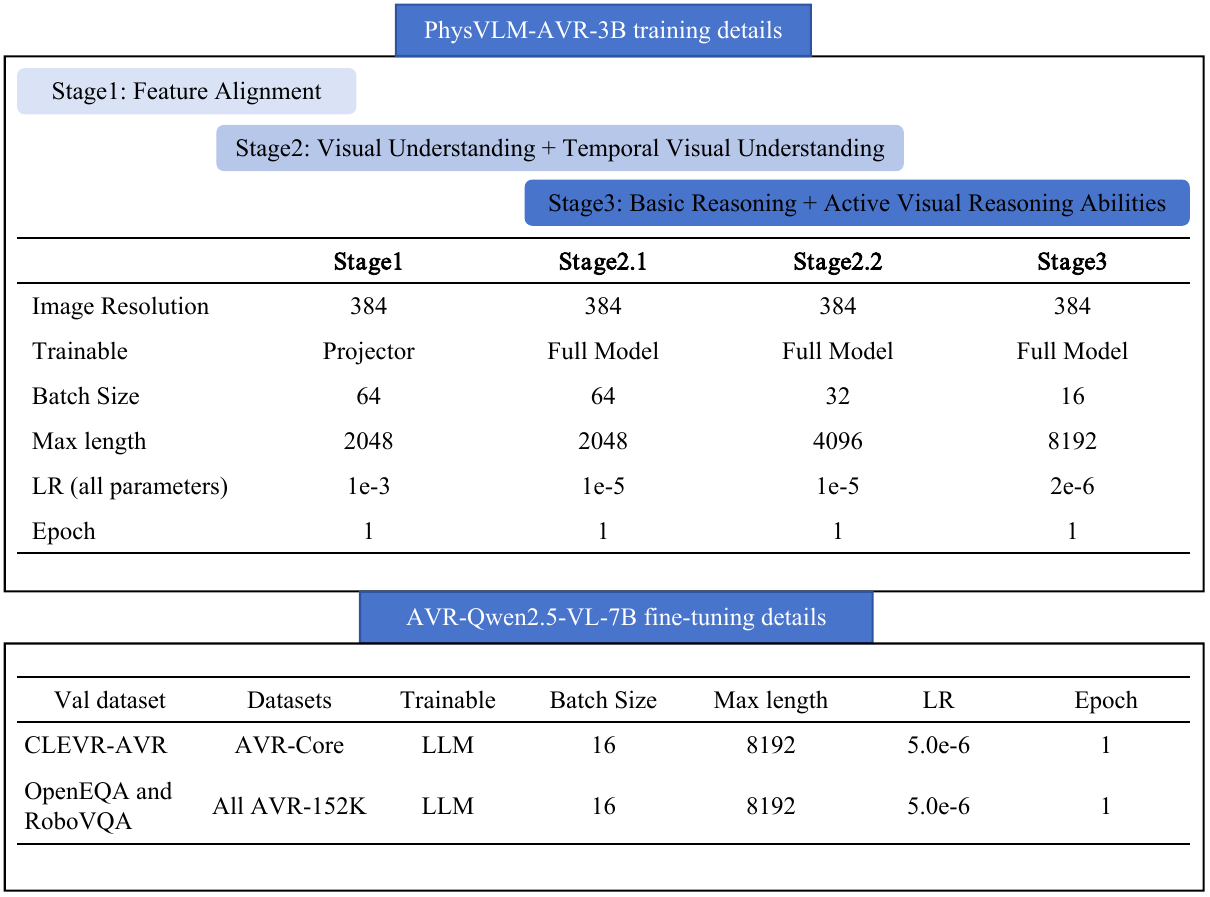}}
    \caption{Training configuration details for PhysVLM-AVR-3B and AVR-Qwen2.5-VL-7B.}
    \label{fig:training_details}
    \end{center}
\end{figure*}

\begin{figure*}[ht]
    \begin{center}
    \centerline{\includegraphics[width=0.7\columnwidth]{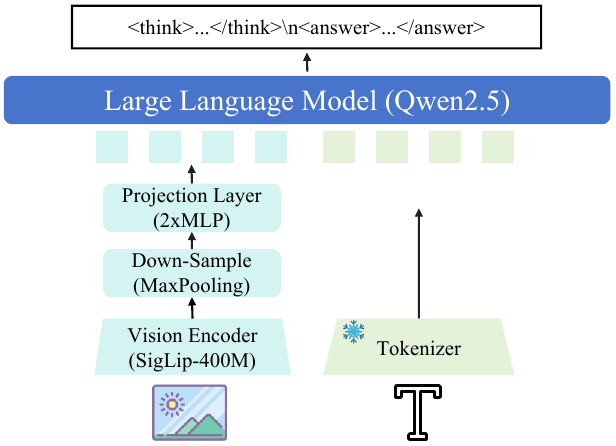}}
    \caption{Model architecture of PhysVLM-AVR.}
    \label{Appendix-model-architecture}
    \end{center}
\end{figure*}

Figure~\ref{fig:training_details} shows the detailed training parameters for PhysVLM-AVR-3B and AVR-Qwen.25-VL-7B, including input image resolution, trainable parameters, batch size, maximum token length, learning rate, and number of epochs. We conducted the training on an Ubuntu server equipped with 8 * NVIDIA A800 GPUs. The main software used was PyTorch=2.6.0, transformers=3.72.0, flash attention2, and DeepSpeed. Figure~\ref{Appendix-model-architecture} showns the architecture of the PhysVLM-AVR.

\section{Baseline Model Input Prompt for CLEVR-AVR Experiments}
\label{Appendix-prompt-input}
In experiments on the CLEVR-AVR benchmark, the inputs for the compared baseline models are prompts that include AVR task instructions and cues for potential actions. The input template is as follows:

\texttt{You are required to perform active visual reasoning: when the information obtained from image observations is insufficient to answer the question, you need to make action decisions to interact with the environment in order to acquire additional visual information relevant to the question. You should continue gathering new observations until you can infer and summarize a reliable answer based on the accumulated visual history. Choose either an answer to the question or an action decision option from the options above. Final option choice follow this format: (your analysis)...<answer>A/B/C/D/E/F...</answer>}

\section{Code Availability}
\label{Appendix-code}

The code for our project is available anonymously at the following link: \url{https://anonymous.4open.science/r/anonymous-je99tt}.

\section{Ethical Considerations and Usage Restrictions}
\label{Appendix-ethical}

Large Language Models (LLMs) and related AI technologies possess the potential for significant societal impacts, both beneficial and detrimental. While they offer capabilities to enhance productivity, creativity, and access to information, it is crucial to acknowledge the inherent risks. These risks include, but are not limited to, the generation and propagation of misinformation, the amplification of existing societal biases, potential for job displacement in certain sectors, and the possibility of misuse for malicious purposes.

The code, data, and models provided in this work are intended strictly for research and development purposes. They must not be employed in any high-risk applications or for activities that could result in harm, discrimination, infringement of rights, or any other negative societal consequences. Users of these resources bear full responsibility for ensuring their applications comply with all applicable laws, ethical guidelines, and responsible AI practices. We explicitly disclaim liability for any misuse of our code, data, or models. We encourage a cautious and ethical approach to the development and deployment of AI technologies.

\clearpage
%%%%%%%%%%%%%%%%%%%%%%%%%%%%%%%%%%%%%%%%%%%%%%%%%%%%%%%%%%%%
\newpage

\section*{NeurIPS Paper Checklist}

\begin{enumerate}

\item {\bf Claims}
    \item[] Question: Do the main claims made in the abstract and introduction accurately reflect the paper's contributions and scope?
    \item[] Answer: \answerYes{} % Replace by \answerYes{}, \answerNo{}, or \answerNA{}.
    \item[] Justification: We reflect the contribution and scope of the paper at the end of the abstract and introduction.

\item {\bf Limitations}
    \item[] Question: Does the paper discuss the limitations of the work performed by the authors?
    \item[] Answer: \answerYes{} % Replace by \answerYes{}, \answerNo{}, or \answerNA{}.
    \item[] Justification: We discuss the limitations of this work in Appendix.

\item {\bf Theory assumptions and proofs}
    \item[] Question: For each theoretical result, does the paper provide the full set of assumptions and a complete (and correct) proof?
    \item[] Answer: \answerNA{} % Replace by \answerYes{}, \answerNo{}, or \answerNA{}.
    \item[] Justification: This article does not involve theoretical results that require proof of assumptions and reasoning.

    \item {\bf Experimental result reproducibility}
    \item[] Question: Does the paper fully disclose all the information needed to reproduce the main experimental results of the paper to the extent that it affects the main claims and/or conclusions of the paper (regardless of whether the code and data are provided or not)?
    \item[] Answer: \answerYes{} % Replace by \answerYes{}, \answerNo{}, or \answerNA{}.
    \item[] Justification: We disclose all the information needed to reproduce the main experimental results of the paper in Experiment, Appendix.

\item {\bf Open access to data and code}
    \item[] Question: Does the paper provide open access to the data and code, with sufficient instructions to faithfully reproduce the main experimental results, as described in supplemental material?
    \item[] Answer: \answerYes{} % Replace by \answerYes{}, \answerNo{}, or \answerNA{}.
    \item[] Justification: We use anonymous code links that follow submission guidelines.

\item {\bf Experimental setting/details}
    \item[] Question: Does the paper specify all the training and test details (e.g., data splits, hyperparameters, how they were chosen, type of optimizer, etc.) necessary to understand the results?
    \item[] Answer: \answerYes{} % Replace by \answerYes{}, \answerNo{}, or \answerNA{}.
    \item[] Justification: We provide information on data splitting in Experiment and Appendix.

\item {\bf Experiment statistical significance}
    \item[] Question: Does the paper report error bars suitably and correctly defined or other appropriate information about the statistical significance of the experiments?
    \item[] Answer: \answerYes{} % Replace by \answerYes{}, \answerNo{}, or \answerNA{}.
    \item[] Justification: The experiments in the article were averaged multiple times, and the random seeds in the experiments were set to 42.

\item {\bf Experiments compute resources}
    \item[] Question: For each experiment, does the paper provide sufficient information on the computer resources (type of compute workers, memory, time of execution) needed to reproduce the experiments?
    \item[] Answer: \answerYes{} % Replace by \answerYes{}, \answerNo{}, or \answerNA{}.
    \item[] Justification: We provide sufficient information in Experiment, Appendix.
    
\item {\bf Code of ethics}
    \item[] Question: Does the research conducted in the paper conform, in every respect, with the NeurIPS Code of Ethics \url{https://neurips.cc/public/EthicsGuidelines}?
    \item[] Answer: \answerYes{} % Replace by \answerYes{}, \answerNo{}, or \answerNA{}.
    \item[] Justification: The research conducted in the paper complied with the NeurIPS Code of Ethics in all respects.

\item {\bf Broader impacts}
    \item[] Question: Does the paper discuss both potential positive societal impacts and negative societal impacts of the work performed?
    \item[] Answer: \answerYes{} % Replace by \answerYes{}, \answerNo{}, or \answerNA{}.
    \item[] Justification: We discuss this in Appendix.
    
\item {\bf Safeguards}
    \item[] Question: Does the paper describe safeguards that have been put in place for responsible release of data or models that have a high risk for misuse (e.g., pretrained language models, image generators, or scraped datasets)?
    \item[] Answer: \answerYes{} % Replace by \answerYes{}, \answerNo{}, or \answerNA{}.
    \item[] Justification: We discuss this in Appendix.

\item {\bf Licenses for existing assets}
    \item[] Question: Are the creators or original owners of assets (e.g., code, data, models), used in the paper, properly credited and are the license and terms of use explicitly mentioned and properly respected?
    \item[] Answer: \answerYes{} % Replace by \answerYes{}, \answerNo{}, or \answerNA{}.
    \item[] Justification: We reference the relevant code and models respecting the license.

\item {\bf New assets}
    \item[] Question: Are new assets introduced in the paper well documented and is the documentation provided alongside the assets?
    \item[] Answer: \answerNA{} % Replace by \answerYes{}, \answerNo{}, or \answerNA{}.
    \item[] Justification: This paper don't release new assets.

\item {\bf Crowdsourcing and research with human subjects}
    \item[] Question: For crowdsourcing experiments and research with human subjects, does the paper include the full text of instructions given to participants and screenshots, if applicable, as well as details about compensation (if any)? 
    \item[] Answer: \answerNA{} % Replace by \answerYes{}, \answerNo{}, or \answerNA{}.
    \item[] Justification: This article does not involve crowdsourcing experiments and studies with humans as subjects.

\item {\bf Institutional review board (IRB) approvals or equivalent for research with human subjects}
    \item[] Question: Does the paper describe potential risks incurred by study participants, whether such risks were disclosed to the subjects, and whether Institutional Review Board (IRB) approvals (or an equivalent approval/review based on the requirements of your country or institution) were obtained?
    \item[] Answer: \answerNA{} % Replace by \answerYes{}, \answerNo{}, or \answerNA{}.
    \item[] Justification: Not applicable to this study.

\item {\bf Declaration of LLM usage}
    \item[] Question: Does the paper describe the usage of LLMs if it is an important, original, or non-standard component of the core methods in this research? Note that if the LLM is used only for writing, editing, or formatting purposes and does not impact the core methodology, scientific rigorousness, or originality of the research, declaration is not required.
    %this research? 
    \item[] Answer: \answerNA{} % Replace by \answerYes{}, \answerNo{}, or \answerNA{}.
    \item[] Justification: The core method development in this research does not involve LLMs as any important, original, or non-standard components.

\end{enumerate}

\end{document}